\documentclass[11pt]{article} 

\usepackage{hyperref}

\usepackage{amsmath}
\usepackage{amssymb}

\usepackage{amsthm}
\newtheorem{theorem}{Theorem}[section]
\newtheorem{proposition}{Proposition}[section]

\usepackage{booktabs}
\usepackage{multirow}
\usepackage{makecell}

\usepackage{siunitx}
\usepackage[boxruled,vlined]{algorithm2e}

\usepackage{bm}
\usepackage{subcaption}

\usepackage{pgfplots}
\pgfplotsset{compat=1.18} 

\usepackage{natbib}
 \bibpunct[, ]{(}{)}{,}{a}{}{,}%

\usepackage{soul}

\title{A Novel Computational Framework for Causal Inference: Tree-Based Discretization with ILP-Based Matching}

\author{
Tianyu Yang\\
Department of Mechanical and Industrial Engineering\\
Northeastern University, Boston, MA 02115, USA\\
\texttt{yang.tianyu@northeastern.edu}
\and
Md. Noor-E-Alam\\
Department of Mechanical and Industrial Engineering\\
Northeastern University, Boston, MA 02115, USA\\
\texttt{mnalam@northeastern.edu}
}

\date{}

\begin{document}

\maketitle

\begin{abstract}
Causal inference is essential for data-driven decision-making, as it aims to uncover causal relationships from observational data. However, identifying causality remains challenging due to the potential for confounding and the distinction between correlation and causation. While recent advances in causal machine learning and matching algorithms have improved estimation accuracy, these methods often face trade-offs between interpretability and computational efficiency. This paper proposes a novel approach that combines a tree-based discretization technique, tailored for causal inference, with an integer linear programming-based matching algorithm. The discretization ensures approximately linear relationships for control datasets within strata, enabling effective matching, while the optimization framework optimizes for global balance. The resulting algorithm yields computational efficiency and less biased ATT estimates compared to state-of-the-art algorithms. Empirical evaluations demonstrate the proposed method's practical advantages over existing techniques in causal inference scenarios.
\end{abstract}


\section{Introduction} \label{sec:intro}

Causal inference is a widely used approach for understanding the impact of interventions, policies, and decisions, and is essential for drawing valid causal conclusions \citep{yao2021survey}. Unlike purely predictive modeling techniques, causal inference aims to uncover cause-and-effect relationships from observational data across various domains, such as economics, healthcare, and the social sciences \citep{spirtes2010introduction}. This enables robust and reliable decision-making that goes beyond identifying mere correlations. 

Theoretically, the most reliable approach for achieving an unbiased estimation of the causal quantities  in causal inference is the Randomized Controlled Trial (RCT) \citep{nitsch2006limits, kovesdy2012observational}. 
RCTs eliminate confounding biases by randomly assigning subjects to treatment and control groups, ensuring balance in both observed and unobserved covariates. However, RCTs are often impractical in observational studies due to high costs in time and money, as well as ethical concerns \citep{nitsch2006limits}. Causal inference methods therefore leverage some advanced techniques to address this concern, as well as improve estimation accuracy and enhance generalizability.

In recent years, causal inference methods have increasingly focused on two major approaches: causal machine learning algorithms and matching algorithms \citep{yao2021survey}. Causal machine learning methods focus on predicting the counterfactual effects, i.e., the unobserved counterfactual outcomes for observations in the treatment group to estimate the average treatment effect on the treated (ATT) \citep{spirtes2010introduction}. Despite their effectiveness, these methods are often criticized for their lack of interpretability \citep{moraffah2020causal}. Evaluating their performance is also challenging, as counterfactual outcome of treatment units are derived from predictive models, and hypothesis tests between the two groups are not applicable. This limitation reduces their practicality in real-world decision-making contexts \citep{moraffah2020causal}. 

In contrast, matching algorithms, which are designed to balance covariates between the selected treated and control groups, are generally more preferable in real world applications since they are interpretable. By selecting specific treated and control groups from the raw dataset, matching allows for an evaluation of balance using common test statistics and p-values. As long as the covariate balance is achieved, matching algorithm can be approximately regarded as a simulation of RCTs and thus will achieve unbiased estimation 
of causal quantity \citep{belitser2011measuring}. Although powerful, most matching algorithms suffer from the curse of dimensionality when estimating treatment effects, which limits their practical applicability in real-world situations. While matching methods find applications across diverse domains including record linkage and case-control studies \citep{christen2012data, austin2011introduction}, our focus is on the causal inference setting where specific methodological requirements—particularly achieving covariate balance—are essential for unbiased treatment effect estimation.

One of the most widely used approaches to overcome the challenges of these two methods is to leverage the propensity score of the dataset. The propensity score is defined as the probability of an observation receiving the treatment, and can be estimated using regression models on the treatment variable \citep{belitser2011measuring}. By mapping all features in the dataset to a single scalar value, the propensity score significantly reduces the computational cost of matching. This technique is known as the Propensity Score Matching (PSM) algorithm. However, many critics argue that PSM 
may not accurately estimate the true ATT, as balancing on the propensity score does not guarantee balance across the full feature space. Moreover, PSM shows higher bias than other matching algorithm in many synthetic datasets \citep{king2019propensity}.

In this paper, we propose a novel interpretable and computationally feasible matching algorithm that leverages machine learning techniques for discretization and optimization techniques for lower bias of ATT estimation compared to the state-of-the-art algorithms. 
Our proposed framework is computationally efficient to the large-scale datasets. The results from our proposed framework also show a small bias among the synthetic datasets.

The rest of the paper is structured as follows: Section 2 presents related literature. Section 3 details our proposed tree-based discretization techniques inspired by the M5 model tree. Section 4 discusses different matching strategies and motivates the development of our ILP formulation. In Section 5, we formally introduce our proposed matching algorithm as an integer linear programming problem. Section 6 provides an overview for the proposed framework. Section 7 presents experimental results on synthetic datasets using widely adopted causal inference metrics. Section 8 demonstrates the practicality of our proposed framework by applying it to a large-scale real-world dataset. Finally, Section 9 concludes the paper.

\section{Literature Review} \label{sec:lit}

The primary goal of causal inference through observational studies is to generate unbiased estimates of causal effects. To achieve this, three key assumptions are raised based on the potential outcome frameworks \citep{yao2021survey}. First, the Stable Unit Treatment Value Assumption (SUTVA), which states that the treatment status of any unit does not affect the potential outcomes of other units, and for each unit, there are no different forms or versions of each treatment level, which lead to different potential outcomes. Secondly, the conditional ignorability assumption, meaning that the outcome $Y$ is independent of the treatment $T$ given covariates $X$. Third, the positivity assumption, which ensures that for any value of $X$, the probability of receiving treatment is strictly between 0 and 1. 

Matching algorithms, which aim to equate the balance between the treatment and control groups, are widely used techniques in real-world causal inference studies. The simplest matching method is exact matching, which pairs treatment and control individuals with identical covariate values \citep{stuart2010matching}. However, this approach becomes impractical when plenty of covariates are involved. Therefore, nearest-neighbor matching (NNM) are proposed to approximate the exact matching process. However, it often fails to achieve a satisfactory covariate balance, leading to biased ATT estimation. Coarsened exact matching (CEM) offers an alternative approach to approximate exact matching by stratifying data into multiple bins and estimating causal effects using weighted averages \citep{blackwell2009cem}. It is effective when the outcome model is highly nonlinear in the confounders since it is a non-parametric method \citep{iacus2012causal}. However, CEM's bins grow exponentially with increasing dimensionality, making it challenging to find exact matches for every treated unit in high-dimensional settings. Other non-parametric models, such as Balance Optimization Subset Selection (BOSS) \citep{nikolaev2013balance} are proposed to approximate covariate balance for each bin. However, BOSS utilizes simulated annealing to achieve results, which is computationally intensive, making it impractical for large-scale datasets.

Another approach to achieving covariate balance is distance-based matching. This method typically involves learning a distance matrix over the covariates to prioritize balance on more important variables through reduced weighted imbalance and thus achive lower bias in ATT estimation. A typical method is Mahalanobis distance matching (MDM), which defines closeness using the Mahalanobis distance matrix \citep{gu1993comparison}. A variant of this method is Genetic Matching \citep{diamond2013genetic}, which uses a generalized Mahalanobis distance while adaptively assigning weights to covariates. However, both methods require large amount of time to calculate matrix inversion and therefore not scalable to the high dimensional datasets. Other distance learning methods, such as matching after learning to stretch (MALTS) \citep{parikh2022malts} and fast large-scale almost matching exactly (FLAME) \citep{wang2021flame}, provide alternative solutions. MALTS learns the distance matrix by assigning large penalties to the imbalance of important covariates. However, it still faces computational challenges in learning distance matrices for large-scale datasets. FLAME utilizes backward feature selection and optimizes based on a combination of two factors. While efficient in computing the distance matrix, its estimation of the causal quantity is highly sensitive to the chosen discretization strategies. It is also recommended to apply FLAME when the features are composed of only categorical features \citep{wang2021flame}. 

Many studies have proposed propensity-score-based approaches \citep{stuart2010matching}. The propensity score represents the probability of a unit receiving treatment. The problem to balance the covariates is then transformed to balance the propensity score in the datasets. This method reduces the dimensionality of the covariates to a single scalar, mitigating the curse of dimensionality in matching. However, critics argue that balancing on the propensity score does not fully 
replicate a blocked randomized experimental design, potentially leading to biased causal estimates \citep{king2019propensity}. This occurs because the propensity score collapses $p$-dimensional covariates into a single scalar: balancing this scalar does not guarantee that individual covariates are balanced, as different covariate combinations can yield identical propensity scores.

One of the most effective strategies for maintaining both interpretability and computational efficiency in matching algorithms is the use of discretization techniques. For example, coarsened exact matching (CEM) \citep{blackwell2009cem} applies equal-width or user-defined rules to discretize the dataset and form coarsened bins. Similarly, balance optimization subset selection (BOSS) \citep{nikolaev2013balance} uses grid-based discretization of the feature space to reduce computational burden. Despite the promise of these approaches in improving the scalability of matching algorithms, relatively few studies have focused on discretization techniques. A possible reason is that discretization may introduce bias—for instance, through arbitrary boundaries that violate outcome continuity or within-stratum heterogeneity—and the extent of this bias can vary depending on the specific technique employed. We provide a detailed discussion of these mechanisms and how our method addresses them in Section \ref{sec:ds} and \ref{sec:overview}.

Although matching algorithms are highly interpretable and often provide accurate estimates of causal quantities, most of them are computationally intensive and suffer from the curse of dimensionality. As a result, many researchers turn to predictive modeling techniques to estimate causal effects more efficiently. For example, random forests have demonstrated strong performance across many datasets. Bayesian Additive Regression Trees (BART), which construct an ensemble of shallow-split trees to predict counterfactual outcomes, are particularly powerful for achieving unbiased ATT estimation in high-dimensional, nonlinear settings. Furthermore, meta-learners such as the X-Learner and R-Learner have been proposed to integrate predictive modeling into the estimation of counterfactual effects for individual observations. Despite their effectiveness, these methods are often criticized for their lack of interpretability, making decision makers hesitate to apply them in practical situations.

There are several methods available to assess the quality of matching between the treatment and control groups produced by matching algorithms. The absolute standardized mean difference (SMD) \citep{belitser2011measuring, zhang2019balance, lin2023balancing, nikolaev2013balance} standardizes the mean difference between matched groups through dividing it by the pooled variance \citep{zhang2019balance}. Typically, SMD values below 0.1 across all dimensions indicate good balance \citep{lin2023balancing}. Variance Ratio (VR) \citep{belitser2011measuring} measures the ratio of variances between the two groups, with values closer to 1 indicating better balance. Many studies use a threshold of 0.5 to 2 across all dimensions as the criterion for balance \citep{rubin2001using, stuart2010matching}. Kolmogorov–Smirnov (K-S) Distance \citep{belitser2011measuring, nikolaev2013balance} quantifies the maximum vertical distance between the cumulative distribution functions (CDFs) of the matched groups, where smaller values indicate greater similarity. Lastly, the Overlap Coefficient (OVL) \citep{belitser2011measuring, lin2023balancing} measures the overlap between two distributions, with values closer to 1 indicating better balance. Although these metrics indicate the balance between the treatment and control groups, they should be viewed only as reference measures. This is because, in most cases, different features contribute unequally to the outcome and thus vary in importance. 

We address the challenges above with several novel aspects that distinguish our approach from existing methods. Existing optimization-based matching methods differ fundamentally in problem formulation and optimization strategy. Table 2 provides a systematic comparison of the state-of-the-art matching algorithms.

\begin{table}[htbp]
\centering
\small
\caption{Comparison of Optimization-Based Matching Methods}
\label{tbl:method_comparison}
\resizebox{\textwidth}{!}
{\begin{tabular}{|l|c|c|c|c|}
\hline
\textbf{Method} & \textbf{Optimization Type} & \textbf{Problem Formulation} & \textbf{Solution Quality} & \textbf{Scalability} \\
\hline
BOSS & Simulated annealing & Stratum-level balance & Heuristic approximation & Limited \\
\hline
Genetic Matching & Evolutionary search & Distance matrix learning & Local optimum & Medium \\
\hline
MALTS & Gradient descent & Distance metric learning & Local optimum & Medium\\
\hline
\textbf{Proposed Approach} & \textbf{ILP} & \textbf{Individual-level balance} & \textbf{Exact per unit} & \textbf{High} \\
\hline
\end{tabular}}

\end{table}

Our contribution lies in three aspects. First, we formulate matching at the individual treatment unit level with explicit optimization of standardized mean difference and variance ratio—the standard balance metrics in causal inference—rather than using surrogate distance metrics. Second, we provide exact linearization of these balance metrics for integer programming, enabling optimal solutions for each treatment unit via commercial ILP solvers rather than relying on heuristic search. Third, we systematically integrate this optimization framework with adaptive tree-based discretization that ensures approximate linearity within strata (Theorem \ref{thm:residual_bound}), exploiting local structure to improve both matching quality and computational efficiency. While we solve a separate ILP for each treatment unit rather than jointly optimizing across all units, this design prioritizes scalability while ensuring each unit achieves optimal balance within its candidate set. Empirically, our method demonstrates superior performance compared to state-of-the-art approaches (Section \ref{sec:syn_res}) while maintaining computational tractability on datasets with hundreds of thousands of observations (Section \ref{sec:cdc_diab}).

The proposed framework has the following key contributions:

\begin{enumerate}

\item \textbf{Causal Tree-Based Discretization with Theoretical Guarantees}. 
We develop a control-based causal M5 tree designed specifically for ATT estimation under treatment imbalance. 
By restricting splits to the control group and adopting an adjusted $R^2$–based stopping rule, 
the resulting leaf nodes approximately satisfy a local linearity condition (Theorem~1). 
This structural property ensures that within-leaf outcome heterogeneity is controlled, 
facilitating statistically coherent matching while maintaining computational scalability for high-dimensional datasets.

\item \textbf{Individual-Level ILP Matching with Direct Balance Optimization}. 
We propose an integer programming formulation that directly optimizes standardized mean difference (SMD) and variance ratio at the individual treatment-unit level. 
Through exact linearization of a hierarchical objective, each matching problem is solved to optimality per treatment unit using standard ILP solvers. 
This formulation provides transparent balance control and enables personalized control selection for ATT (and IATT) estimation, 
distinguishing it from distance-learning or heuristic-based matching approaches.

\item \textbf{Integrated Framework with Empirical Validation}. 
We integrate tree-based discretization and optimization-based matching into a unified framework in which each component is theoretically motivated and empirically validated. 
Ablation studies isolate the contribution of the ILP component, and extensive synthetic and real-data experiments demonstrate stable and competitive ATT estimation across diverse data-generating scenarios.

\end{enumerate}

\section{Discretization Techniques} \label{sec:ds}

Discretization is the process of partitioning a continuous 
or high-dimensional feature space into a finite number of discrete strata. 
In causal inference, discretization addresses the curse of dimensionality 
in exact matching: With $p$ covariates, the number of possible covariate patterns grows exponentially, quickly rendering exact matching computationally infeasible even for moderate $p$. For example, discretizing age from a continuous 
variable into strata such as $[18,30)$, $[30,50)$, $[50,80]$ enables 
tractable matching within each stratum while approximating the conditions 
of a randomized experiment.

However, discretization introduces a fundamental trade-off 
between tractability and bias. It can introduce bias through three 
mechanisms: (1) \emph{arbitrary boundaries} may violate continuity of 
covariate-outcome relationships (e.g., treating individuals aged 29 and 31 
as dissimilar); (2) \emph{within-stratum heterogeneity} occurs when 
observations with different outcomes are grouped together; and (3) 
\emph{loss of information} arises from collapsing continuous variation. 
The extent of this bias depends critically on the discretization strategy 
employed.

Although widely used in causal inference algorithms, existing discretization strategies exhibit distinct 
limitations (Table~\ref{tab:discretization_comparison}). Equal-width 
binning creates arbitrary boundaries. Grid-based approaches (CEM, BOSS) 
suffer exponential growth ($O(k^p)$ strata), limiting scalability. Quantile 
methods may inappropriately split regions where outcomes vary smoothly. 
K-means clustering ignores outcome information. Entropy-based methods are 
prone to overfitting. Our tree-based method addresses these limitations by 
adaptively partitioning based on outcome prediction quality ($R^2_{\text{adj}}$), 
ensuring within-stratum homogeneity at computational efficiency $O(pn\log n)$. 
Critically, we employ a single decision tree rather than an ensemble, 
preserving interpretability through explicit, human-readable split rules. 
As Theorem~1 demonstrates (Section~3.2), this approach ensures approximate 
linearity within leaves, essential for unbiased treatment effect estimation 
when combined with our optimization-based matching procedure (Section~5).

\begin{table}[htbp]
\centering
\small
\caption{Comparison of Discretization Techniques for Matching Algorithms}
\resizebox{\textwidth}{!}{
\begin{tabular}{|l|l|l|l|}
\hline
\textbf{Method} & \textbf{Approach} & \textbf{Complexity} & \textbf{Limitations} \\
\hline
Equal-width & Equal-sized intervals & $O(n)$ & Arbitrary boundaries; unequal bin sizes \\
\hline
Quantile & Equal sample sizes & $O(n \log n)$ & May split smooth regions \\
\hline
K-means & Distance-based clustering & $O(nkp)$ & Ignores outcome; expensive \\
\hline
Grid (CEM, BOSS) & Cartesian product & $O(k^p)$ & Exponential growth \\
\hline
Entropy & Minimize entropy & $O(n^2 p)$ & Overfitting; small bins \\
\hline
\textbf{Tree (ours)} & \textbf{Outcome-based} & \textbf{$O(pn \log n)$} & \textbf{Assumes local linearity} \\
\hline
\end{tabular}}
\label{tab:discretization_comparison}
\end{table}

\subsection{Causal M5 Model Tree} \label{sec:c5m}

M5 model tree is a tree-based structure which is designed to generate leaf nodes where the linearity assumption holds \citep{pal2009m5}. The linearity assumption is a powerful property for estimating ATT. Suppose the dataset $D=\{T,X,Y\}$, where $T$ refers to the binary treatment variable and is $n\times1$, $T=1$ if this unit gets treated, and $T=0$ if this unit does not receive treatment. $X$ refers to the feature space and is $n\times p$, and $Y$ refers to the outcome variable that we concern and is $n\times 1$. With linear assumption $Y=\alpha T+X\beta+\epsilon$, then the unbiased ATT estimation from potential outcome framework would be exactly $E(Y(T=1)|T=1)-E(Y(T=0)|T=1)=\alpha$. Additionally, the local linearity assumption at each node aligns with the theory of heterogeneous treatment effects. This assumption not only has the potential to reduce the computational cost of the matching algorithm but also enhances its interpretability. However, to the best of our knowledge, M5 model tree is rarely used in causal inference applications. One reason the researcher may hesitate to implement the M5 model tree is that its key assumption often does not hold when applied directly to the original dataset. Specifically, treatment and control units will not generally fall under the same linear model unless the treatment variable $T$ is uncorrelated with all covariates in $X$, or the true ATT within the leaf node is exactly zero. To explain this, let us consider leaf node $O_l, l\in\{1, 2,...,l\}$, it contains both control units and treatment units to estimate ATT (otherwise the positivity assumption discussed in \ref{sec:lit} will be violated). Then we have the following proposition and proof.

\begin{proposition} [Linearity Within a Node] \label{prop:node_lin}
    Within a node, the treatment and control units will not fall into the same linear model unless T is uncorrelated with all covariates in X or the true ATT for this node is 0.
\end{proposition}

The proof for Proposition \ref{prop:node_lin} is presented in the \textbf{Online Supplement Appendix S1 Appendix A}.

Proposition \ref{prop:node_lin} provides theoretical motivation for our control-based tree construction. This design choice reflects an inductive bias that prioritizes stability under severe treatment imbalance, 
at the cost of potentially missing effect modifiers that do not influence baseline outcomes. Practical considerations further support this design choice. In many observational studies, the number of treated units is substantially smaller than that of controls. If tree splits were determined using treated units, even a small number of splits could result in leaf nodes containing too few treated observations, leading to unstable estimation, inflated variance, or violations of the positivity assumption.

To address this challenge, we adapt the M5 model tree to the causal inference setting by proposing a causal M5 model tree, in which the tree structure is learned exclusively from the control subset $D^0 \subset D$. Specifically, splitting rules are determined using only control units, and linear models are fitted within each leaf under the assumption that sufficient control data are available to support reliable estimation. Treatment units are subsequently assigned to the learned leaves for treatment effect estimation, but they do not influence the tree construction process. Under the conditional ignorability assumption $(Y(0), Y(1)) \perp T \mid X$, the control outcome $Y(0)$ encodes confounding relationships between covariates and potential outcomes. Our control-based splitting criterion, which maximizes standard deviation reduction in $Y(0)$, naturally identifies covariates associated with confounding. The subsequent ILP optimization (Section \ref{sec:ilp}) explicitly enforces balance on these covariates within each leaf, ensuring conditions for unbiased ATT estimation.  Empirical evidence further clarifies the operating regime of this design choice. 
As shown in \textbf{Online Supplement S1 Appendix H.2}, this inductive bias tends to be advantageous in settings where treated and control units exhibit covariate distribution shift, while under covariate alignment the two constructions perform comparably.

This design prioritizes structural stability and statistical reliability under severe treatment imbalance. While the tree is constructed without direct access to treatment outcomes, this does not preclude the discovery of heterogeneous treatment effects. When treatment effect heterogeneity is associated with covariates that also predict the control outcome, different leaves naturally exhibit different treatment effects. We empirically demonstrate this behavior in Section \ref{sec:cdc_diab}. Importantly, this approach introduces an explicit inductive bias: it assumes that relevant treatment effect heterogeneity is aligned with variations in the control outcome surface. Under this assumption, restricting splits to the control group enables robust and interpretable tree structures while avoiding unreliable partitioning driven by sparse treated samples.

The split rule of our causal M5 model tree is based on standard deviation reduction (SDR), defined as follows:

\begin{equation} \label{equ:SDR}
    SDR = \sigma(D^0)-\bigg(\frac{|D^0_1|}{|D^0|}\cdot\sigma(D^0_1)+\frac{|D^0_2|}{|D^0|}\cdot\sigma(D^0_2)\bigg)
\end{equation}

Here, $D^0$ denotes the subset of the control group that reaches the parent node, while $D^0_1$ and $D^0_2$ represent the corresponding subsets that fall into the two child nodes after the split. The function $\sigma(\cdot)$ computes the standard deviation of the target variable within a node, and $|\cdot|$ denotes the number of samples in the corresponding subset. The splitting rule facilitates efficient partitioning of the dataset by grouping data points with similar target values into the same child node. By maximizing the reduction in standard deviation, the split encourages the formation of more homogeneous subsets, which not only speeds up the tree construction process but also improves the accuracy of the local linear models fitted at the leaves.

Different from the original M5 model tree, we decide whether we should split further for the tree based on the $R_{adj}^2$, which is defined as below:

\begin{equation}
    R^2_{adj}=1-(1-R^2)\cdot\frac{|D^0|-1}{|D^0|-p-1}
\end{equation}

Here, $p$ refers to the number of features in the feature space. The $R^2$ is computed by fitting a linear regression model to the control data within each node. We define $S_1$ and $S_2$ as binary indicators for whether the left and right child nodes, respectively, meet the splitting criterion based on $R^2_{\text{adj}}$ improvement. Consequently, the decision to further split the tree is decided by the following rule:

\begin{align} 
    \label{equ:s_improve_1}
    & S_1 = I(R^2_{adj}(D^0_1) - R^2_{adj}(D^0) - \lambda I(\theta-|D^0_1|>0) > 0) \\
    & S_2 = I(R^2_{adj}(D^0_2) - R^2_{adj}(D^0) - \lambda I(\theta-|D^0_2|>0) > 0) \\
    & S = S_1 \lor S_2
    \label{equ:s_improve_2}
\end{align}

Here, $\theta$ is a hyperparameter representing the minimum number of control units required in a child node to maintain reliable linear model fitting. To ensure adequate degrees of freedom while guarding against overfitting, we set $\theta = \max(30, 2p)$, where $p$ is the number of features. This adaptive criterion ensures each leaf contains sufficient observations relative to dimensionality, maintaining a minimum sample-to-parameter ratio of 2:1. For datasets with $p \leq 20$ (typical in our experiments), this yields $\theta \leq 40$, providing $(n_l - p - 1) \geq p$ degrees of freedom for reliable $R^2_{\text{adj}}$ estimation. Critically, while $\theta$ determines leaf size for establishing linearity, the computational complexity of subsequent matching is controlled independently by the candidate set size $\psi$ in Algorithm 2, ensuring scalability regardless of $\theta$. $S$ is a boolean value indicating whether the split should be performed for this node. A split is allowed when $S=1$, which means, at least one child node satisfies the adjusted $R^2$ improvement criterion after penalization. Since the adjusted $R^2$ indicates the linearity of the control model, the causal M5 model tree can efficiently address the local interpretability and enables further subgroup analysis.

\begin{algorithm}[htbp]
\small
\caption{Discretization Through Causal M5 Model Tree}
\label{alg:disc}

\SetKwInput{KwInput}{Input}
\SetKwInput{KwOutput}{Output}
\SetKwFunction{Split}{Split}
\SetKwProg{Fn}{Function}{}{end}

\KwInput{Dataset $D$ with $p$ features, $\lambda \leftarrow 0.1$, $\theta \leftarrow \max(30,2p)$}
\KwOutput{Fully developed binary tree $Tree$}

Initialize $Tree$ and normalize dataset $D$\;
$D^0 \leftarrow D(T=0)$ \tcp{Extract control dataset from original dataset}
\Split{$Tree, D^0$}\;

\Fn{\Split{$Tree, D^0$}}{

    $SDR \leftarrow -\infty$\;
    $SP \leftarrow \mathrm{None}$\;
    $\eta \leftarrow -\infty$\;

    $R^2_{prt} \leftarrow R^2$ from regressing $Y$ on $X$ using $D^0$\;
    $(R^2_{prt})_{adj} \leftarrow 1-(1-R^2_{prt})\cdot\frac{|D^0|-1}{|D^0|-p-1}$\;

    \For{$j=1,2,\ldots,p$}{
        $S_j \leftarrow$ unique set of values of feature $j$\;
        
        \For{$s_j \in S_j$}{
            $(D^0_1)_{tmp} \leftarrow$ left child node split by $X_j \le s_j$\;
            $(D^0_2)_{tmp} \leftarrow$ right child node split by $X_j > s_j$\;

            $SDR_{tmp} \leftarrow \sigma(D^0)-
            \left(
            \frac{|(D^0_1)_{tmp}|}{|D^0|}\sigma((D^0_1)_{tmp})
            +
            \frac{|(D^0_2)_{tmp}|}{|D^0|}\sigma((D^0_2)_{tmp})
            \right)$\;

            \If{$SDR_{tmp}>SDR$}{
                $SDR \leftarrow SDR_{tmp}$\;
                $SP \leftarrow X_j$\;
                $\eta \leftarrow s_j$\;
            }
        }
    }

    $D^0_1 \leftarrow$ left child node split by $SP \le \eta$\;
    $D^0_2 \leftarrow$ right child node split by $SP > \eta$\;

    $R^2_{lc} \leftarrow R^2$ from regressing $Y$ on $X$ using $D^0_1$\;
    $R^2_{rc} \leftarrow R^2$ from regressing $Y$ on $X$ using $D^0_2$\;

    $(R^2_{lc})_{adj} \leftarrow 1-(1-R^2_{lc})\cdot\frac{|D^0_1|-1}{|D^0_1|-p-1}$\;
    $(R^2_{rc})_{adj} \leftarrow 1-(1-R^2_{rc})\cdot\frac{|D^0_2|-1}{|D^0_2|-p-1}$\;

    \If{$(R^2_{lc})_{adj}-\lambda I(\theta-|D^0_1|>0)>(R^2_{prt})_{adj}$ 
    \textbf{or}
    $(R^2_{rc})_{adj}-\lambda I(\theta-|D^0_2|>0)>(R^2_{prt})_{adj}$}{
        $Tree \leftarrow$ add split point and threshold $(SP,\eta)$ to $Tree$\;
        \Split{$Tree,D^0_1$}\;
        \Split{$Tree,D^0_2$}\;
    }
    \Else{
        \Return{$Tree$}\;
    }
}
\Return{$Tree$}\;

\end{algorithm}

\subsection{Theoretical Justification for Linearity Within Leaf Nodes
} 
\label{sec:lin_assump}

The causal M5 model tree creates leaf nodes where control data approximately satisfy linearity through iterative splitting. Consider leaf node $O_l$ containing control data $D_l^0$ with $|D_l^0| = n_l$ observations. Our splitting criterion selects the split maximizing standard deviation reduction and proceeds only if at least one child node achieves $R^2_{\text{adj}}$ improvement over the parent node (Equations \ref{equ:s_improve_1} to \ref{equ:s_improve_2}). When the splitting process terminates at a leaf, we can bound the residual variance relative to the total variance.

\begin{theorem}[Residual Bound Within Leaves]
\label{thm:residual_bound}
Let $Y_i$ denote the outcome for observation $i$ in leaf node $O_l$ with control data $D_l^0$, and let $\hat{Y}_i = \beta_0 + X_i^T \beta$ denote the prediction from the fitted linear model. If the adjusted R-squared satisfies $R^2_{\text{adj}}(D_l^0) = \zeta$, then the mean squared residual satisfies
\begin{equation}
\frac{1}{n_l} \sum_{i \in D_l^0} (Y_i - \hat{Y}_i)^2 \leq (1 - \zeta) \cdot \frac{n_l - 1}{n_l - p - 1} \cdot \text{Var}(Y \mid D_l^0).
\end{equation}
\end{theorem}

The proof of Theorem \ref{thm:residual_bound} is provided in \textbf{Online Supplement S1 Appendix B}.

This theorem quantifies the quality of linear approximation within each leaf while providing built-in protection against overfitting. The factor $(n_l - 1)/(n_l - p - 1)$ plays a dual role: it accounts for degrees of freedom in the residual bound, and it reflects how adjusted R-squared guards against spurious fit. With the selection of $\theta=\max(30, 2p)$, leaves with high $R^2_{\text{adj}}$ provide stronger evidence of meaningful linear relationships rather than overfitting.

\subsection{Limitations of the Causal M5 Tree Model}

While our splitting criterion promotes linearity within leaves, certain scenarios may challenge this assumption. First, the method requires sufficient sample size relative to dimensionality. The stopping criterion $\theta = \max(30, 2p)$ necessitates adequate control sample size to support multiple leaf nodes. When $N_c / p$ is small, the tree may produce too few leaves to capture outcome heterogeneity effectively. Second, sharp discontinuities in the outcome function may resist linear approximation, although the tree typically splits near such boundaries, creating leaves where the outcome is approximately constant. Third, datasets exhibiting complex global nonlinearity may not benefit from a locally linear approach. Fourth, control-based construction may be less effective when treatment effect heterogeneity is driven by \emph{pure effect modifiers}—covariates that modify $\tau(X)$ but have little influence on the baseline outcome $Y(0)$. Because splits are determined using control outcomes, such covariates may not trigger partitioning, potentially pooling units with heterogeneous treatment effects within the same leaf. In these settings, more flexible ensemble methods may be preferable, albeit at the cost of interpretability.

\section{Matching Strategies and Robust ATT Estimation} \label{sec:1:k}

In this section, we explore the relationship between different matching strategies (1:1, 1:k, k:k) for estimating the ATT. Building on these insights, we then introduce our proposed integer linear programming method in Section \ref{sec:ilp}.

In this section, we discuss the compatibility of three matching strategies with stratification techniques, including tree-based discretization: (1) 1:1 matching, where each treated unit is matched to a single control unit; (2) 1:k matching, where each treated unit is matched to multiple control units; and (3) k:k matching, where treated units are divided into one or more groups, and each group is matched to a corresponding group of control units. Stratification techniques apply specific rules to partition the data into multiple strata. The within-stratum covariate balance is assumed to hold.

We present an example of a stratum in Table \ref{tbl:rob_ex}. In the table, $Y^T$ represents the outcome value for treatment units, while $Y^C$ represents the outcome value for control units. For example, the first row indicates that the outcome value for observation $T_1$ is 1 and outcome value for observation $C_1$ is 0.

\begin{table}[htbp] 
\centering
\caption{Example for robust ATT estimation within a stratum. $Y^T$ represents the outcome value for treatment units, while $Y^C$ represents the outcome value for control units}
\label{tbl:rob_ex}
\resizebox{0.2\textwidth}{!}{
\begin{tabular}{|c|c|c|c|}
\hline
$T$ & $Y^T$ & $Y^C$ & $C$ \\ \hline
$T_1$ & 1 & 0 & $C_1$ \\ \hline
$T_2$ & 1 & 0 & $C_2$ \\ \hline
$T_3$ & 0 & 0 & $C_3$\\ \hline
$T_4$ & 0 & 0 & $C_4$\\ \hline
$T_5$ & 0 & 0 & $C_5$\\ \hline
    &   & 1 & $C_6$\\ \hline
    &   & 1 & $C_7$\\ \hline
\end{tabular}
}
\end{table}

Assume we are using 1:1 matching, the robust ATT estimation can be achieved if and only if the maximum discordant pairs are matched within the stratum \citep{islam2019robust}. Therefore, the matched groups $\Omega_{1:1}$=\{($T_1$, $C_1$), ($T_2$, $C_2$), ($T_3$, $C_6$), ($T_4$, $C_7$)\}. The mean of the ATT estimation will be $ATT_{1:1}=mean(ATT_{\Omega\{1,1\}})=0$. 

We now consider the robust matched pairs for 1:k matching, which gives $\Omega_{1:k}$=\{($T_1$, $C_1\sim C_7$), ($T_2$, $C_1\sim C_7$), ($T_3$, $C_1\sim C_7$), ($T_4$, $C_1\sim C_7$), ($T_5$, $C_1\sim C_7$)\}, the robust ATT then can be calculated accordingly, $ATT_{1:k}=((1-2/7)\cdot2+(0-2/7)\cdot3)/5=0.114$. If we consider k:k matching, then $\Omega_{k:k}$=\{($T_1\sim T_5$, $C_1\sim C_7$)\}. The robust ATT can be calculated through $ATT_{k:k}=2/5-2/7=0.114$. 

The example provides an interesting insight: robust ATT estimation is not necessarily equivalent between the three matching strategies (1:1, 1:k, and k:k matching) within the stratum. This raises an important question: which approach yields unbiased ATT estimates?

We argue that unbiased estimation is achieved by 1:k matching and k:k matching. When applying a stratum-based technique, covariate balance is assumed within each stratum, meaning that balance holds at the group level rather than for individual units. Consider $C_1 \sim C_7$ as seven independent observations with identical covariate values; their observered outcomes should be consistent (either all 0 or all 1). Otherwise, this would violate the Stable Unit Treatment Value Assumption (SUTVA) discussed in Section \ref{sec:lit}. Compared to 1:1 matching, 1:k and k:k matching align with the group-level balance assumption, making them more appropriate for unbiased ATT estimation in stratified settings.

\begin{figure} [htbp]
    \centering
    \includegraphics[width=\textwidth]{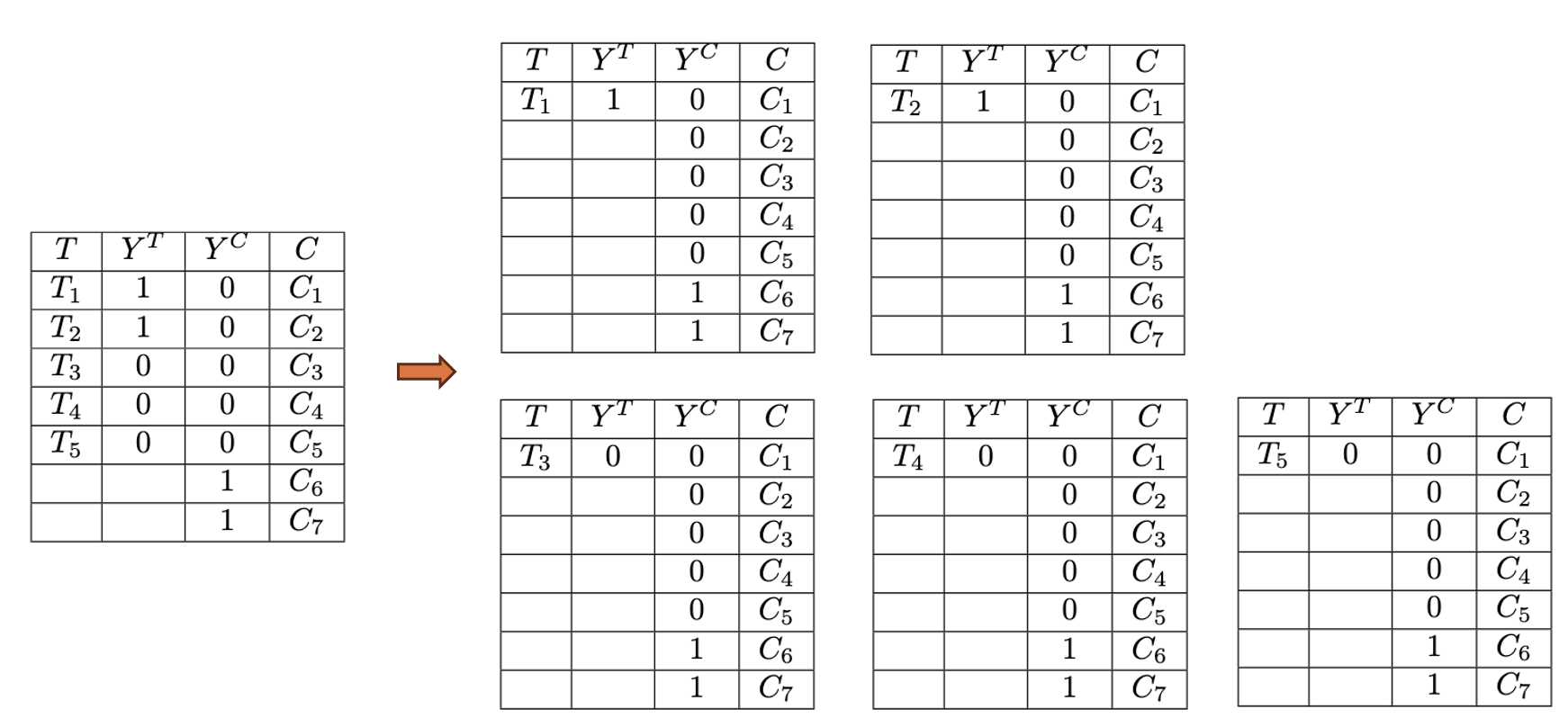}
    \caption{Example of stratum-level equivalence for 1:k matching and k:k matching, the overall average ATT is the average ATT estimation for each decomposed stratum.}
    \label{fig:decomp}
\end{figure}

The discussion above rules out 1:1 matching under stratum-based matching algorithms, as it introduces unexpected bias in the estimation of causal quantities.

One may question whether 1:k matching and k:k matching always yield the same robust estimation of ATT. To address this, we provide the following theorem along with its proof:

\begin{theorem} [Stratum-Based Matching Equivalence] \label{the:strat}
    For stratum-based matching methods, the robust ATT estimation within each stratum remains the same for both 1:k matching and k:k matching.
\end{theorem}



The proof for Theorem \ref {the:strat} is presented in \textbf{Online Supplement S1 Appendix B}. Theorem \ref{the:strat} also leads to an interesting proposition below:

\begin{proposition} [Stratum-level equivalence for ATT Estimation] \label{prop:strat_decompose}
    Assume a dataset utilize stratification techniques and generate $B$ strata. For stratum $b, b=\{1,2,...,B\}$, there are $n_t^b$ treatment units and $n_c^b$ control units, the following two methods to calculate the overall ATT is equivalent:

    1. Calculate ATT directly by the weighted average of ATT estimation for all stratum, with the weights defined by $w_b =n_t^b/\sum_{b=1}^B{n_t^b}$. Assume the ATT estimation for stratum $b$ is $\tau^{b}$, then 
    \begin{equation}
        ATT =\sum_{b=1}^Bw_b\tau^{b}=\frac{1}{\sum_{b=1}^B{n_t^b}}\sum_{b=1}^Bn_t^b\cdot\tau^{b}.
    \end{equation}

    2. Decompose each stratum $b$ into $n_t^b$ strata. The decomposed stratum $\kappa, \kappa=\{1, 2,...,n_t^b\}$ contains $1$ treatment unit and $n_c$ control units. The individual-level ATT (IATT) can be calculated accordingly, denote as $\tau^{\kappa}$, the overall ATT can be calculated by
    
    \begin{equation}
        ATT=\frac{1}{\sum_{b=1}^B{n_t^b}}\cdot\sum_{b=1}^B\sum_{\kappa=1}^{n_t^b}\tau^{\kappa}
    \end{equation}
\end{proposition}

Proposition \ref{prop:strat_decompose} provides an alternative interpretation of matching within a stratum. For example, we can interpret Table \ref{tbl:rob_ex} as 5 strata of 1:k matching stratum in Figure \ref{fig:decomp}.

Proposition \ref{prop:strat_decompose} also indicates that 1:k matching is more flexible than k:k matching as it allows unit-level stratification. For example, in Table \ref{tbl:rob_ex}, k:k matching must assign $T_1 \sim T_5$ to $C_1 \sim C_7$. However, it probably happens that, $C_8$ is spatially close to $T_2$ but not the others, then we could add $C_8$ in the decomposed stratum in $T_2$ only. This makes 1:k matching more flexible than k:k matching. This strategy can alleviate the bias introduced by stratification in many stratification-based k:k matching algorithms and will lead to an ATT estimation with lower bias.

It is worth noting that 1:1 matching and 1:k matching are suitable for calculating individual ATT (IATT), whereas k:k matching is not. The ability to compute IATT is crucial in certain causal inference applications. For example, estimating the individual treatment effect for each treated patient can support personalized treatment recommendations. This consideration further motivates our framework’s use of 1:k matching.

\section{Integer Linear Programming Construction for 1:k Matching} \label{sec:ilp}

Before presenting our integer linear programming formulation, we establish notation used throughout this section. Table \ref{tab:ilp_notation} defines all indices, decision variables, and parameters in our optimization model. Throughout this section, $\kappa$ indexes strata associated with single treatment units, $i$ indexes control units within a stratum, and $j$ indexes features.

\begin{table}[htbp]
\centering
\small
\caption{Notation for Integer Linear Programming Formulation}
\begin{tabular}{|c|l|l|}
\hline
\textbf{Symbol} & \textbf{Type} & \textbf{Definition} \\
\hline
\multicolumn{3}{|c|}{\textit{Indices}} \\
\hline
$\kappa$ & Index & Decomposed strata associated with single treatment \\

& & unit, $\kappa \in \{1, \ldots, N_t\}$ \\ 
$i$ & Index & Control unit in stratum $\kappa$, $i \in \{1, \ldots, n_c^\kappa\}$ \\
$j$ & Index & Feature (covariate), $j \in \{1, \ldots, p\}$ \\
\hline
\multicolumn{3}{|c|}{\textit{Decision Variables}} \\
\hline
$g_i$ & Binary & Equals 1 if control unit $i$ is selected, 0 otherwise \\
$\epsilon$ & Continuous & Maximum weighted mean difference across features \\
$a$ & Continuous & Maximum weighted deviation of selected controls \\
\hline
\multicolumn{3}{|c|}{\textit{Parameters}} \\
\hline
$x_{ij}^\kappa$ & Continuous & Value of feature $j$ for control unit $i$ in stratum $\kappa$ \\
$\mu_j^\kappa$ & Continuous & Value of feature $j$ for the treatment unit in $\kappa$ \\
$w_j$ & Continuous & Weight for feature $j$ (from OLS regression) \\
$M_1$ & Constant & Constraint enforcement constant (set to $10^6$) \\
$M_2$ & Constant & Objective prioritization constant (any value larger than \\ & & $10 \cdot p \cdot \max_j w_j$, here we set it to be $10^6$) \\
\hline
\end{tabular}
\label{tab:ilp_notation}
\end{table}

Proposition \ref{prop:strat_decompose} in Section \ref{sec:1:k} indicates that a large stratum can be interpreted by multiple smaller strata, each containing a single treatment unit and multiple control units. While this approach mitigates the between-strata bias, the within-stratum bias remains unsolved. For example, $T_1$ may ideally be matched only to $C_1$ and $C_3$, whereas matching $T_1$ to $C_1 \sim C_7$ could introduce a relatively larger bias. Therefore, in this section, we leverage the integer programming approach to efficiently eliminate within-strata bias and achieve smaller bias of ATT estimation.

\subsection{SMD and Variance Ratio for 1:k Matching} \label{sec:smd_vr}

Before introducing our proposed integer programming framework, we first present two widely used metrics for assessing matching quality, as our proposed matching framework is constructed leveraging these metrics to achieve covariate balance. The first metric is called standardized mean difference (SMD), SMD measures the difference in means between the treatment and control groups. The other metric is called variance ratio, which quantifies the difference in variance between the two groups. Ideally, if the two groups are perfectly matched, then the absolute SMD should be 0, and variance ratio should be 1. 

For continuous features, the absolute SMD and variance ratio is defined below:

\begin{equation} \label{equ:smd}
    |SMD| = \frac{|\mu_1 - \mu_0|}{\sqrt{(\sigma_1^2+\sigma_0^2)/2}}
\end{equation}
\begin{equation} \label{equ:vr}
    VR = \sigma_1^2/\sigma_0^2
\end{equation}

Here, $\sigma_1^2$ refers to the variance of the treatment group for the target feature, $\sigma_0^2$ refers to the variance of the control group for the target feature.

In Equation~\ref{equ:vr}, since each decomposed stratum contains only one treatment unit, $\sigma^2_1 = 0$. Therefore, to achieve balance (VR $\approx$ 1), $\sigma^2_0$ should be minimized.

However, in Equation~\ref{equ:smd}, when $\sigma^2_1 = 0$ and $\sigma^2_0 \approx 0$, the denominator $\sqrt{(\sigma^2_1 + \sigma^2_0)/2}$ becomes extremely \emph{small} (approaching zero). Consequently, even a small numerator $|\mu_1 - \mu_0|$ would result in an extremely large SMD. Therefore, it is \emph{critical} to minimize $|\mu_1 - \mu_0|$ to prevent SMD from becoming unbounded. This is why minimizing the mean difference 
$|\mu_1 - \mu_0|$ must take priority over minimizing variance $\sigma^2_0$ in our objective function formulation.

We provide a simple numerical example to illustrate the characteristics of a well-balanced control group. Suppose we are balancing only one feature $\mathcal{X}$ within stratum $b$. The stratum includes one treatment unit $\mathcal{X}^b_t=5$ and five control units $\mathcal{X}^b_c=\{3, 4, 4.5, 6, 7\}$. When apply 1:k matching, one might consider several choices of control groups: 
$(\Omega^b_0)_1=\{4.5\}, (\Omega^b_0)_2=\{3, 4, 6, 7\}, (\Omega^b_0)_3=\{3, 7\}, (\Omega^b_0)_4=\{4, 6\}$. In this case, we believe $(\Omega^b_0)_4$ is the best choice of set of control units. Specifically, the mean of $(\Omega^b_0)_4$ is 5, while selecting additional units in $(\Omega^b_0)_2$ would introduce greater variability. 

\subsection{Constraints of the Integer Program and Linearization} \label{sec:const}

Based on the previous discussion, we introduce the construction of ILP to minimize the SMD and variance ratio within each stratum. Our optimization problem is hierarchical in nature. We first enforce covariate balance between treated and control units, and only refine the solution based on secondary criteria among well-balanced candidates.

Assume the datasets include $\mathcal{N}_t$ treatment units, and $\mathcal{N}_c$ control units. For stratum $\kappa$, $\kappa\in(1,\mathcal{N}_t)$, it includes one treatment unit $t^\kappa$ and $n_0^\kappa$ control units $\{c^\kappa_{1}, c^\kappa_{2},..., c^\kappa_{n_0^\kappa}\}$, $n_0^\kappa<\mathcal{N}_c$. Although $n_0^\kappa$ control units are included in stratum $\kappa$, we are supposed to select a subset of these control units to minimize the within-stratum balance. Therefore, we call the group of control units \textbf{candidate control units} for $t^\kappa$.

To indicate which control units in the candidate control group is selected for the matched control group, we define indicators $g_i, i\in(1,n_0^\kappa)$. If $g_i=1$, then $c^\kappa_{i}$ is selected in the matched control group; if $g_i=0$, then control unit $c^\kappa_{i}$ is not selected for the control group.

We consider balancing across $p$ features, indexed by $j \in \{1, \ldots, p\}$. For treatment unit $t^\kappa$, the value of feature $j$ is $\mu^\kappa_j$, while for control units $c^\kappa_i$, $i \in \{1, \ldots, n^\kappa_c\}$, the value is $x^\kappa_{ij}$. The weight for feature $j$ given by the OLS estimation from linear regression is $w_j$, then the weighted distance between $c^\kappa_i$ and $t^\kappa$ for feature $j$ is given by $w_j|x^\kappa_{ij} - \mu^\kappa_j|$. Since $g_i$ indicates whether this control unit $c^\kappa_i$ is selected, the absolute mean difference between the $T_\kappa$ and the matched control group would be $\sum_{i=0}^{n_0}g_iw_j|x^\kappa_{ij}-\mu_j^\kappa|$. Clearly, minimizing this value should be part of the objective function in order to achieve a close-to-zero SMD within the stratum. Now considering variability of the selected control units, the maximum weighted deviation across all features is $\max_{i,j} \{g_i w_j |x^\kappa_{ij} - \mu^\kappa_j|\}$. This value should be minimized as well to reduce $\sigma_0^2$ in Eq.\ref{equ:vr}. As discussed in Section \ref{sec:smd_vr}, the absolute SMD is more powerful than variance ratio. We assume $M_2$ is a large enough value defined in Table \ref{tab:ilp_notation}, the objective function derived from the theory to minimize balance metrics across all $p$ features should be:

\begin{equation}
\min \max_{i,j} \{g_i w_j |x^\kappa_{ij} - \mu^\kappa_j|\} + 
M_2 \max_{j \in \{1,\ldots,p\}} \left\{ \sum_{i=1}^{n^\kappa_c} g_i w_j |x^\kappa_{ij} - \mu^\kappa_j| \right\}
\label{equ:ori_obj}
\end{equation}

It is obvious that this objective function is too complicated for practical implementation. Now we introduce $\epsilon \geq 0$ to represent the maximum weighted mean difference across all features $j$: 
$\epsilon = \max_{j \in \{1,\ldots,p\}} \sum_{i=1}^{n^\kappa_c} g_i w_j |x^\kappa_{ij} - \mu^\kappa_j|$, which represents the maximum weighted distance between the mean of the matched control group and the treatment unit $t^\kappa$ across all dimensions. Consequently, $\epsilon$ can represent the second term in Eq.\ref{equ:ori_obj}. Similarly, we define $a=argmax_{i,j}\{g_iw_j|x^\kappa_{ij}-\mu_j^\kappa|\}, a\geq0$ as the maximum weighted distance to $t_\kappa$ across all dimensions. To enforce this, we introduce $M_1$ defined in Table \ref{tab:ilp_notation} and the constraint $w_j|x^k_{ij}-\mu_j^\kappa|-M_1(1-g_i)\leq a$.

\begin{enumerate}
    \item if $g_i=1$, then the constraint simplifies to $w_j|x^k_{ij}-\mu_j^\kappa|\leq a$, effectively bounding the maximum deviation within the selected control units. 

    \item If $g_i=0$, then $M_1(1-g_i)$ is large enough to make the constraint trivially satisfied, ensuring that it does not influence the choice of $a$.
\end{enumerate}

By incorporating $\epsilon$ and $a$, we then have the following two constraints: (i) $\sum_{i=1}^{n_0}g_iw_j|x_{ij}^\kappa-\mu_j^\kappa| \leq \epsilon$ (ii) $w_j|x^\kappa_{ij}-\mu_j^\kappa|-M_1(1-g_i) \leq a$. Noticing that the two constraints are nonlinear, we adopt linearization techniques to handle the absolute value terms. Our formulation prioritizes minimizing covariate imbalance($\epsilon$) before minimizing variance ($a$).
The constant $M_2$ enforces this hierarchy, serving as a computational surrogate for lexicographic optimization. Consequently, the final proposed integer linear program (ILP) framework is presented below:

\begin{subequations} \label{equ:ilp}
    \begin{align}
        & \min \quad  a + M_2 \epsilon \label{eq:obj} \\
        & \sum_{i=1}^{n^\kappa_c} g_i w_j (x^\kappa_{ij} - \mu^\kappa_j) \leq \epsilon, 
          \quad \forall j \in \{1,\ldots,p\} \label{eq:eps1} \\
        & \sum_{i=1}^{n^\kappa_c} g_i w_j (\mu^\kappa_j - x^\kappa_{ij}) \leq \epsilon, 
          \quad \forall j \in \{1,\ldots,p\} \label{eq:eps2} \\
        & w_j (x^\kappa_{ij} - \mu^\kappa_j) - M_1(1 - g_i) \leq a, 
          \quad \forall i,j \label{eq:a1} \\
        & w_j (\mu^\kappa_j - x^\kappa_{ij}) - M_1(1 - g_i) \leq a, 
          \quad \forall i,j \label{eq:a2} \\
        & \sum_{i=1}^{n^\kappa_c} g_i \geq 1 \label{eq:pos}
    \end{align}
\end{subequations}

The variance term $a$ refines the solution among groups achieving comparable $\epsilon$: since Theorem~\ref{thm:residual_bound} ensures only approximate (not strict) linearity, minimizing $a$ reduces bias from residual heterogeneity within leaves.

Here, we employ two constants: $M_1 = M_2 = 10^6$. The constant $M_1$ enforces constraint validity for unselected units (Equations~\ref{eq:a1}--\ref{eq:a2}), while $M_2$ enforces hierarchical prioritization in the objective (Equation~\ref{eq:obj}). The choice of $M_2$ is grounded in the following result (proof in \textbf{Online Supplement Appendix S1 }):

\begin{proposition}[Sufficient Condition for Hierarchical Prioritization]
\label{prop:big_m}
Let $\bar a := n_c^\kappa \max_j w_j\,(\max_{\mathrm{feat}}-\min_{\mathrm{feat}})$.
Assume $\epsilon$ is optimized with fixed precision $\Delta>0$.
If $M_2>\bar a/\Delta$, then any feasible solution with smaller $\epsilon$ is preferred regardless of $a$.
\end{proposition}

After normalization to $[0,1]$ with $n^\kappa_c = 20$ and $\max_j w_j \leq 10$, this requires $M_2 > 200$. We set a sufficiently large value of $M_2$ to enforce the hierarchical objective while maintaining computational efficiency relative to lexicographic optimization. We note that Proposition~\ref{prop:big_m} provides a sufficient rather than tight condition, and smaller values of $M_2$ may still perform well in practice.

We also add constraint (\ref{equ:ilp}f) to avoid the solution where all $g_i=0$, ensuring the feasibility of a reasonable solution. This constraint agrees with the positivity assumption of matching. The simplest way to estimate $w_j$ is to conduct a linear regression of the features on the outcome. 

\begin{figure} [htbp]
    \centering
    \subfloat[Graph Interpretation of $\epsilon$ of stratum $\kappa$]{%
        \includegraphics[width=0.44\textwidth]{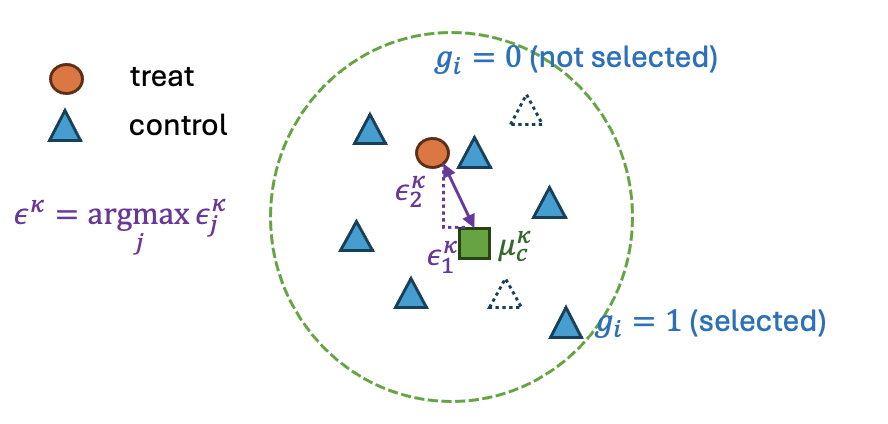}%
    }%
    \hfill
    \subfloat[Graph Interpretation of $a$ of stratum $\kappa$]{%
        \includegraphics[width=0.44\textwidth]{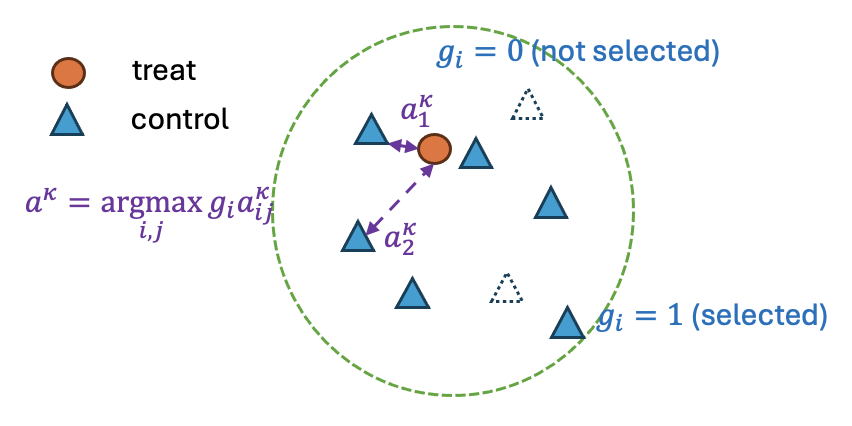}%
    }%
    \caption{The Graph interpretation of $\epsilon$ and $a$ in Eq.\ref{equ:ilp}, the green rectangle denoted by $\mu_c^\kappa$ in (a) refers to the mean of selected control units (control units with $g_i=1$)}
    \label{fig:graph_int}
\end{figure}

We also provide graphical interpretations in Figure \ref{fig:graph_int} to illustrate the role of $\epsilon$ and $a$ in the proposed framework. In Figure (\ref{fig:graph_int}(a)), the distance of the mean of selected control units to $t^\kappa$ is $\{\epsilon_{1}^\kappa, \epsilon_{2}^\kappa, ..., \epsilon_{p}^\kappa\}$ is a $p$ dimensional vector, where each element $\epsilon_j^\kappa$ represents the components of the weighted distance in feature $j$. The scalar $\epsilon^\kappa$ is then computed as $\epsilon^\kappa=argmax_{j}\epsilon^\kappa_{j}$. By minimizing $\epsilon^\kappa$ in Eq.\ref{equ:ilp}, the mean difference of the selected control units and $t^\kappa$ is controlled. In Figure (\ref{fig:graph_int}(b)), denote the difference between each individual selected control unit and $t^\kappa$ to be $g_ia_i^\kappa, i\leq 1\leq n^\kappa_c$. Then $a^\kappa$ represents the largest value among selected control units and across all the dimensions, i.e., $argmax_{i,j}\{g_i a_{i,j}^\kappa\}$. By minimizing $a^\kappa$ in Eq.\ref{equ:ilp}, $t^\kappa$ and selected control groups achieve close-to-zero absolute SMD and low within-stratum variance. By enforcing these conditions, the framework effectively balances covariates while controlling variability, leading to more reliable causal effect estimates.

One notable observation is that, based on our experiments, the normalization of the feature space $X$ will increase the quality of the selected control units. When processing our datasets, we utilized the default \texttt{MinMaxScaler} function from Python package.

By combining the causal M5 model tree discretization technique with stratum-based 1:k ILP matching, we construct matched datasets without relying on predictive models to estimate counterfactual outcomes. We refer to this model-free approach as the M5 causal tree with model-free estimation (\texttt{M5C\_MF}).

\section{Overview for the Proposed Algorithm} \label{sec:overview}

\begin{figure} [htbp]
    \centering
    \resizebox{0.8\textwidth}{!}{\includegraphics[width=\textwidth]{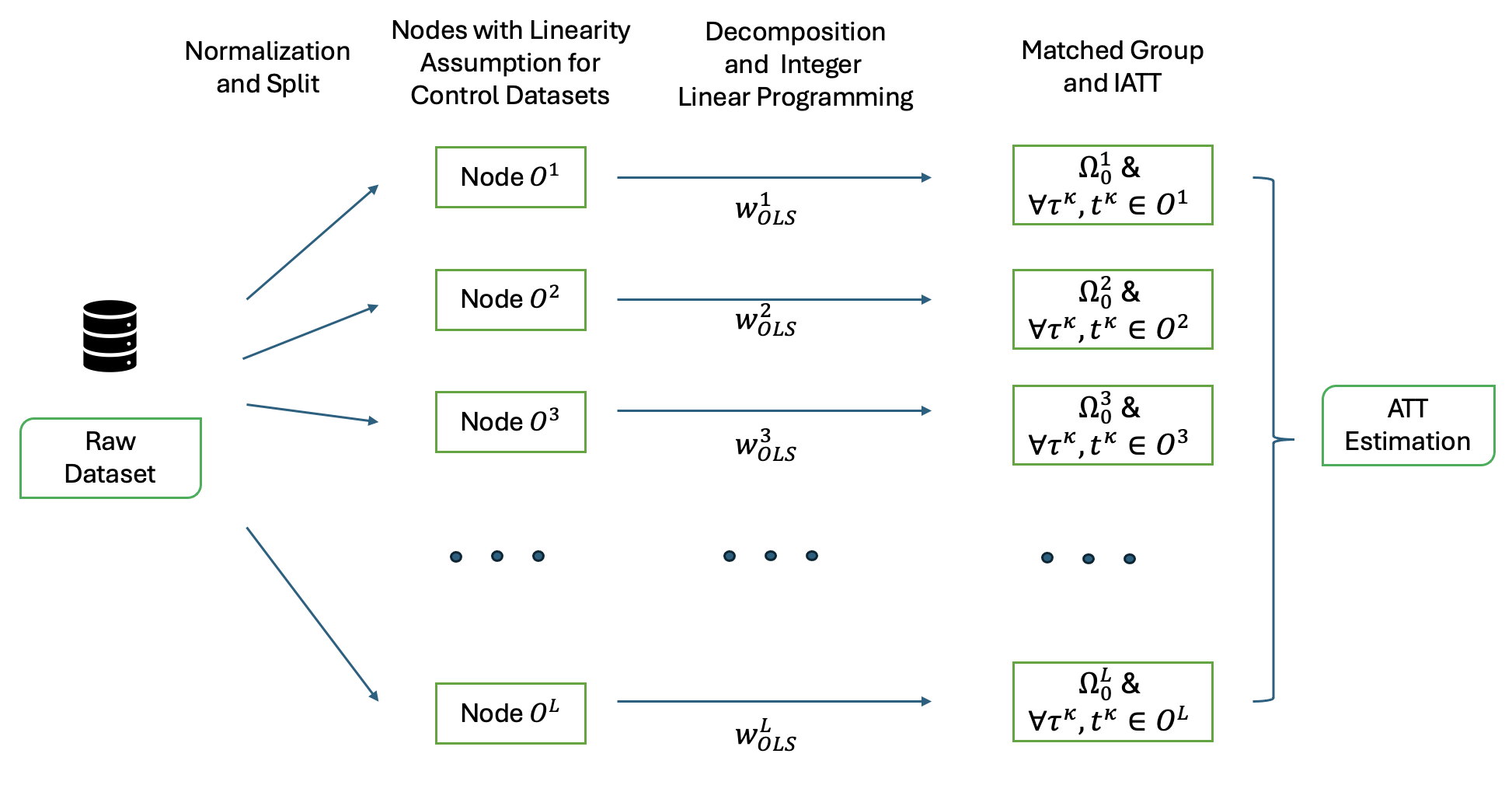}%
    }
    \caption{Flowchart for implementation of the proposed framework}
    \label{fig:flow}
\end{figure}

In this section, we provide an overview of the proposed algorithm. The method first discretizes the dataset using the proposed M5 tree, as described in Section \ref{sec:ds}. Then, for each treatment unit within each leaf node, the ILP approach introduced in Section \ref{sec:ilp} is applied. The overall workflow is illustrated in Figure \ref{fig:flow}.

\subsection{Quality of ILP}

According to our experiments, omitting the discretization step can lead to biased ATT estimation, even if commonly used balance metrics, such as standardized mean difference (SMD), variance ratio, Kolmogorov–Smirnov (K-S) statistic, and overlap appear satisfactory after ILP. One possible explanation for this is the varied importance of each feature across different strata.


\subsection{Interpretability}

\begin{table}[htbp]
\small
\centering
\caption{Comparison of Interpretability Across Causal Inference Methods}
\label{tab:interpretability_comparison}
\small
\begin{tabular}{@{}lp{3cm}p{3.3cm}p{3.8cm}@{}}
\toprule
\textbf{Method} & \textbf{Prediction Mechanism} & \textbf{Interpretability Challenge} & \textbf{Validation Approach} \\
\midrule
BART & Ensemble of trees & Averaging obscures decision logic & Unverifiable counterfactual predictions \\
\addlinespace
Random Forest & Bootstrap aggregation & Black-box predictions & Unverifiable counterfactual predictions \\
\addlinespace
Meta-learners & Nested ML models & Multiple model layers & Model fit diagnostics only \\
\addlinespace
PSM & Propensity score & May not capture full balance & Balance on propensity score only \\
\addlinespace
\textbf{Ours} & \textbf{Single tree plus ILP} & \textbf{Human-readable split rules} & \textbf{Standard metrics on all covariates} \\
\bottomrule
\end{tabular}
\end{table}

Our framework prioritizes interpretability through two design choices that distinguish it from existing methods (Table~\ref{tab:interpretability_comparison}). As discussed in Section~2, causal machine learning methods such as BART, Random Forests, and meta-learners lack interpretability due to ensemble predictions and black-box counterfactual estimation \citep{bassan2025makes}. In contrast, our algorithm uses a single decision tree with explicit split rules, enabling stakeholders to understand stratification logic without machine learning expertise. For instance, rules such as ``if BMI $>$ 30 and Age $<$ 50'' are directly interpretable for healthcare practitioners or policymakers.

Equally important, interpretability requires verifiable validation. Counterfactual prediction methods produce unverifiable estimates, as true counterfactual outcomes are never observed. Our ILP-based matching instead constructs explicit groups from observed data, enabling validation through standard balance diagnostics (SMD, variance ratio, K-S tests). This allows practitioners to audit matching quality, verify that confounding has been addressed, and conduct sensitivity analyses—capabilities essential for justifying causal claims in high-stakes applications. As noted in Section \ref{sec:intro}, this verifiability distinguishes matching algorithms from predictive models and makes our approach accessible to non-technical stakeholders.

\subsection{Addressing Discretization Challenges}

Having introduced all components of our framework—the causal M5 tree (Section~\ref{sec:ds}), matching strategies (Section~\ref{sec:1:k}), and ILP formulation (Section~\ref{sec:ilp})—we now discuss how their combination addresses the three key challenges of discretization identified in Section~\ref{sec:ds}.

First, our tree design eliminates \emph{arbitrary boundaries} through adaptive splitting based on standard deviation reduction (Equation~\ref{equ:SDR}). Unlike fixed schemes (equal-width, quantile) that impose predetermined split points, our approach selects boundaries to maximize outcome homogeneity within child nodes, ensuring split points 
emerge naturally from the data structure and align with genuine outcome variation.

Second, our $R^2_{\text{adj}}$-based stopping criterion (Equations~\ref{equ:s_improve_1}--\ref{equ:s_improve_2}) directly mitigates \emph{within-stratum heterogeneity}. As established in heorem~\ref{thm:residual_bound}, this ensures control data within each leaf approximately satisfy a linearity 
assumption, with mean squared residual bounded by $(1-\tau) \cdot \frac{n_l-1}{n_l-p-1} \cdot \text{Var}(Y|D^0_l)$ when achieving $R^2_{\text{adj}} = \tau$. This prevents grouping heterogeneous observations and enables unbiased ATT estimation (Section~\ref{sec:ilp}).

Third, while discretization inherently involves \emph{information loss}, our framework minimizes its impact through two mechanisms: (1) local linearity (Theorem~\ref{thm:residual_bound}) ensures minimal bias from collapsing continuous features, and (2) individual-level 1:k matching (Theorem~\ref{the:strat}) allows each treated unit to be matched to a tailored subset of controls, preserving more granular information than group-level k:k matching. Table~\ref{tab:discretization_solutions} summarizes our design choices.

\begin{table}[htbp]
\small
\centering
\caption{How Our Framework Addresses Discretization Challenges}
\label{tab:discretization_solutions}
\small
\begin{tabular}{@{}p{3.5cm}p{4.5cm}p{5cm}@{}}
\toprule
\textbf{Challenge} & \textbf{Description} & \textbf{Our Solution} \\
\midrule
Arbitrary boundaries & Fixed discretization schemes impose splits that may 
violate covariate-outcome continuity & Adaptive splitting based on standard 
deviation reduction (Eq.~1); boundaries emerge from data structure \\
\addlinespace
Within-stratum heterogeneity & Observations with different outcomes grouped 
together violate local homogeneity assumption & $R^2_{\text{adj}}$ stopping 
criterion ensures approximate linearity within leaves (Theorem~1) \\
\addlinespace
Information loss & Discretization reduces granularity of continuous 
features & Local linearity reduces bias impact; 1:k matching preserves 
individual-level information (Theorem~2) \\
\bottomrule
\end{tabular}
\end{table}

\subsection{Time Complexity}

In the first stage of the proposed algorithm, we employ a tree-based framework based on the M5 model tree for discretization. Let $n$ denote the number of samples and $p$ the number of features. The overall time complexity of constructing the tree is $\mathcal{O}(p\cdot nlogn)$. 

In the second stage of the proposed algorithm, assume there are $n_t$ treatment units and $n_c$ control units. First we consider $\psi$ nearest control units as the candidate control units, this step takes $\mathcal{O}(n_cp)$. Then we solve an integer linear program using \texttt{Gurobi} solver. Assume the worst case iteration to be $L$, then the time complexity is $\mathcal{O}(L p^3+L\psi p)$. Since we only consider ATT, the overall time compelxity for this step is $O(n_tn_cp+L p^3n_t+L \psi pn_t)$. In our experiment, we set $\psi=20$ and $L=1000$.

\begin{algorithm} [htbp]
    \small
    \caption{Stratum Based Algorithm for 1:k Matching}   
    \SetKwInput{KwInput}{Input}
    \SetKwInput{KwOutput}{Output}
    \label{alg:1_to_k} 
    \KwInput{$M$ $\leftarrow$ 1,000,000, $\psi$ $\leftarrow$ 20, dataset $D$}
    \KwOutput{Estimated Average Treatment Effect on the Treated (ATT)}
    
    \text{$w$ $\leftarrow$ array of OLS coefficients of linear regression on $X$ to $Y$}
        
    \text{Initialize empty list of dataframe $CCU$} \tcp{$CCU$ is the candidate control units set}

    \text{$\mathcal{N}_t \leftarrow$ number of treatment units in dataset $D$}

    \text{$\mathcal{N}_C \leftarrow$ number of control units in dataset $D$}

    \text{$p \leftarrow$ dimension of the features of dataset $D$}

    \For {$\kappa = 1,2,..., \mathcal{N}_t$}{
        \text{Initialize $Dist$ to the size of $\mathcal{N}_C$} 
        \For{$i = 1,2,..., \mathcal{N}_C$}{
            \text{Update $Dist_i \leftarrow \sqrt{\sum_{j=1}^pw_j(t^\kappa_{j}-c^\kappa_{ij})^2}$} \tcp{$j$ refers to the index of the current feature}
        }
        \text{$CCU^\kappa \leftarrow$ select $\psi$ nearest control units based on $Dist$}
        \text{$\Omega_c^\kappa \leftarrow$ output from $Solver(t^\kappa,  CCU^\kappa, M)$ with ILP constructed by Eq.\ref{equ:ilp}}
        
        \If {$\Omega_c^\kappa$ not empty}{
            \text{$CCU^\kappa \leftarrow \Omega_c^\kappa$}
            
            \text{$\tau^\kappa \leftarrow y(t^\kappa)-\frac{1}{|c^\kappa|}\sum_{c^\kappa\in \Omega_c^\kappa}y(c^\kappa)$}\tcp{Individual ATT for treatment unit $t^\kappa$}
        }
    }
    \text{$\tau \leftarrow \frac{1}{|\tau^\kappa|}\sum_{\kappa=1}^{\mathcal{N}_t}\tau^\kappa$}
    \tcp{The overall ATT for dataset $D$}
    \Return $\tau$
\end{algorithm}

By combining the time complexity together, the final time complexity of the proposed algorithm is $\mathcal{O}(pnlogn+n_tn_cp+L p^3n_t+L \psi pn_t)$.

\section{Experimental Results} \label{sec:syn_res}

In this section, we compare the ATT estimation performance of our proposed algorithm with several state-of-the-art methods. 

We evaluated our proposed framework on three synthetic datasets. We present results for one dataset here and reporting the other two in \textbf{Online Supplement S1 Appendix E}.

For our proposed methods, we include two frameworks: (i) M5C\_M, which first applies the propose discretization techniques to the dataset and then predicts the counterfactual outcomes using a linear regression model within each leaf node, we present this method in detail in \textbf{Online Supplement S1 Appendix D}; and (ii) M5C\_MF, which also utilizes the proposed discretization technique, but subsequently use the integer linear programming (ILP) procedure described in Section \ref{sec:ilp} within each node. The proposed algorithms are implemented in Python.

The benchmark algorithms considered include: (iii) Inverse Probability of Treatment Weighting (IPTW), (iv) One-to-One Nearest Neighbor Matching on Propensity Score (1-PSNNM), (v) Bayesian Additive Regression Trees (BART) \citep{hill2011bayesian}, (vi) Random Forests (RF) \citep{wager2018estimation}, (vii) R-learner \citep{zhao2023causal}, and (viii) Genetic Matching \citep{diamond2013genetic}. Genetic Matching is implemented in R, while the other benchmark algorithms are implemented in Python. All experiments are conducted on a 2.4 GHz Quad-Core Intel Core i5 processor.

\textbf{Synthetic Dataset}: Dataset $hyb20var$:

This dataset is generated through the following DGP:

\begin{align*}
&x_{i,1}...,x_{i,5} \overset{\textit{iid}}{\sim} \mathcal{U}(0, 10), p_j\overset{\textit{iid}}{\sim}\mathcal{U}(0.1, 0.9), 
x_{i,j}\overset{\textit{iid}}{\sim}\mathcal{B}(p_j), j = 6, 7,..., 20\\
&\epsilon_i \overset{\textit{iid}}{\sim} \mathcal{U}(0, 1),
\mathcal{N}_t=200,
\mathcal{N}_c=19,800
\\
&y_0 = 0.5x_{i,1} + 0.3x_{i,2} + 0.2x_{i,1}x_{i,6} + 0.5sin(x_{i,3}) +0.5x_{i,4}^2+0.3\mathbb{I}(x_7=1) +  \epsilon_i \\
&y_1 =  0.5x_{i,1} + 0.3x_{i,2} + 0.2x_{i,1}x_{i,6} + 0.5sin(x_{i,3}) +0.5x_{i,4}^2+0.3\mathbb{I}(x_7=1) + 2 + \epsilon_i \\
&\sum\nolimits_{i=1}^{\mathcal{N}_t + \mathcal{N}_c} t_i=\mathcal{N}_t, t_i = \{0, 1\} \\
&y=y_1t_i + y_0(1-t_i)
\end{align*}

In this DGP, we included 20 variables. The first 5 variables are continuous variable, while the rest 15 variables are binary variable. Among 20 variables, 7 variables are used to generate the outcome. We randomly select 200 units out of 20,000 to receive the treatment, with the remaining units serving as controls. The true ATT in this DGP is $2$. We repeat this process for 30 iterations and record the absolute bias in the estimated ATT for each run.

When benchmarking this dataset, we include an additional baseline algorithm, Almost Matching Exactly (FLAME) \citep{wang2021flame}, using the R package \texttt{FLAME}. During preprocessing, we discretize each continuous variable into quantile-based bins. 

\begin{figure} [htbp]
    \centering

    \subfloat[Absolute bias for the ATT estimation]{%
        \resizebox{0.49\textwidth}{!}{
            \includegraphics[width=\textwidth]{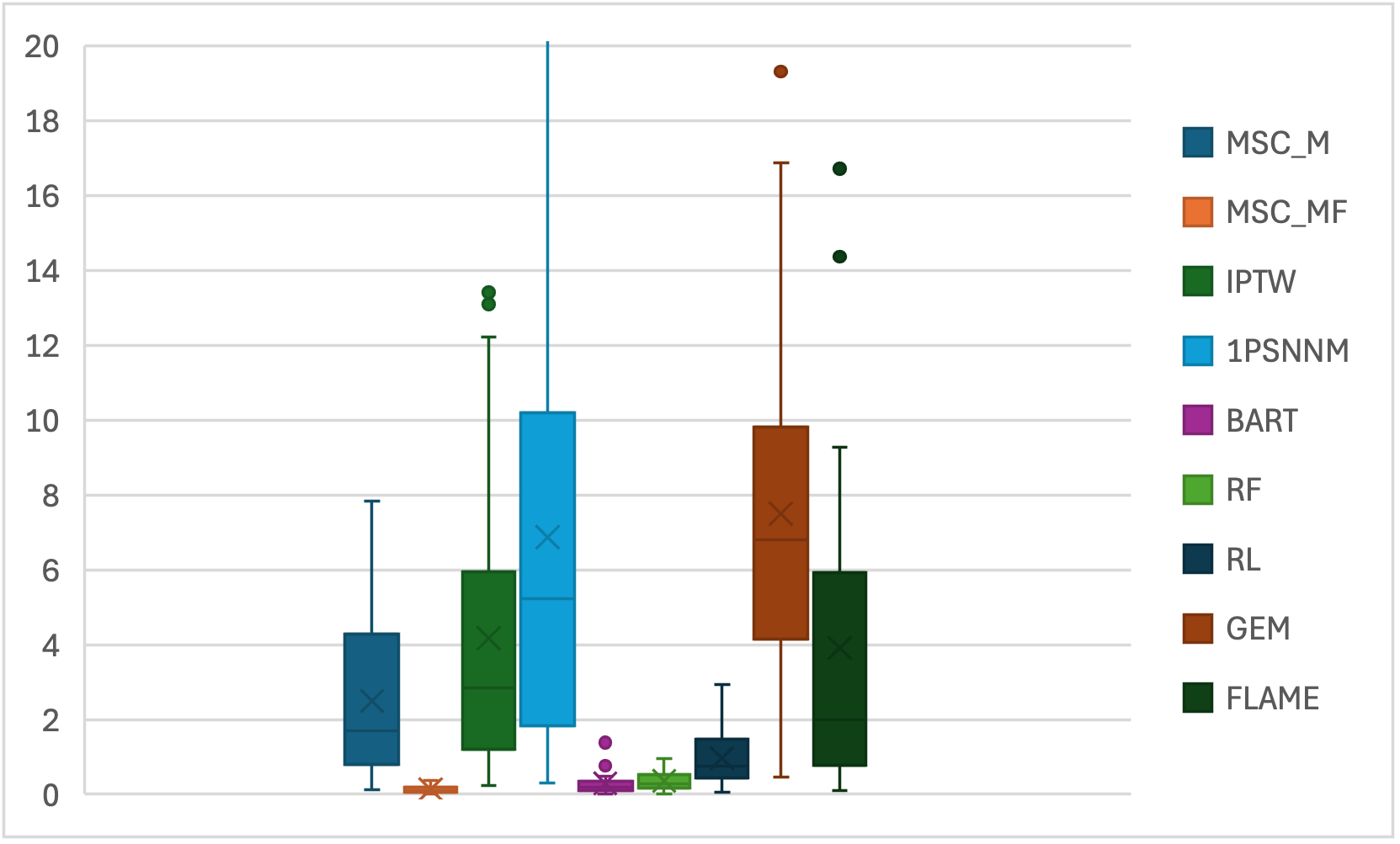}%
        }
    }%
    \hfill
    \subfloat[95\% confidence interval for the absolute bias]{%
        \resizebox{0.49\textwidth}{!}{
        \includegraphics[width=\textwidth]{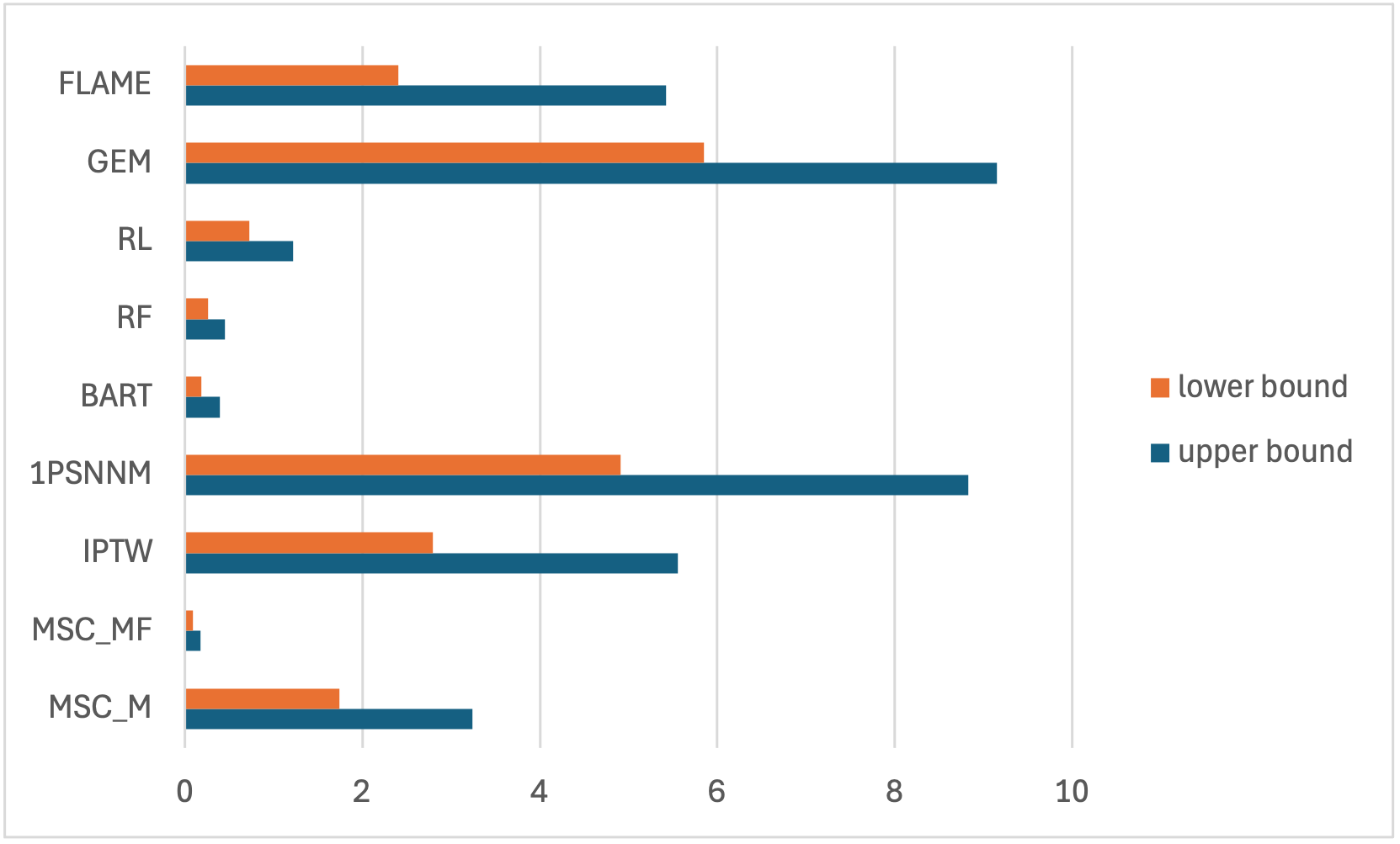}%
        }
    }%
    \hfill
    \caption{Absolute bias to the ATT estimation for $hyb20var$}
    \label{fig:hyb20var}
\end{figure}

We present the absolute bias with respect to the true ATT estimation in Figure \ref{fig:hyb20var}. Specifically, Figure \ref{fig:hyb20var}(a) shows the boxplot of the absolute estimation bias regarding ATT estimation across all runs, while Figure \ref{fig:hyb20var}(b) illustrates the corresponding confidence intervals of the bias to the ATT estimation. Figure \ref{fig:hyb20var} shows that in this case, the performance of  M5C\_MF consistently achieves smaller bias  compared to the other state-of-the-art algorithms. This indicates that M5C\_MF is preferable when robust and accurate estimation of ATT is required.

The results on the three synthetic datasets show that the proposed algorithm achieves low bias while maintaining interpretability. In particular, the M5C\_MF method consistently yields lower bias and demonstrates robustness in terms of confidence interval.

\section{Case Study: CDC Diabetes Dataset} \label{sec:cdc_diab}

The CDC Diabetes Health Indicators Dataset is a public health dataset provided by the U.S. Centers for Disease Control and Prevention (CDC) and hosted by the UCI Machine Learning Repository. It originates from the 2015 Behavioral Risk Factor Surveillance System (BRFSS), the largest ongoing telephone-based health survey system in the world. The dataset includes responses from 253,680 adults in the United States and covers 21 health-related indicators. Among these, 2 are continuous variables and the remaining are categorical. After applying one-hot encoding to the categorical variables, the dataset contains a total of 27 features. The primary objective of analyzing this dataset is to explore the prevalence and correlates of type 2 diabetes, as well as general health behaviors within the U.S. adult population.

In this case study, the outcome variable is \texttt{Diabetes\_binary}, where a value of 1 indicates that the individual has been diagnosed with diabetes, and 0 otherwise. We consider the variable \texttt{GenHlth} as the treatment. Notably, \texttt{GenHlth} receives the highest absolute coefficient in a logistic regression model predicting \texttt{Diabetes\_binary}, suggesting a strong association between general health status and diabetes diagnosis. \texttt{GenHlth} represents the participants’ self-evaluation of their general health status. A value of 5 indicates that the individual considers themselves to be in ``poor" health. Therefore, we define the treatment variable as 1 for participants whose \texttt{GenHlth} value is 5 (indicating poor health), and 0 for the rest values. The dataset includes 12,081 adults with a \texttt{GenHlth} value of 5 and 241,599 adults with a \texttt{GenHlth} value other than 5.

\begin{figure} [htbp]
    \centering

    \subfloat[ATT estimation for each algorithm]{%
        \resizebox{0.49\textwidth}{!}{
            \includegraphics[width=\textwidth]{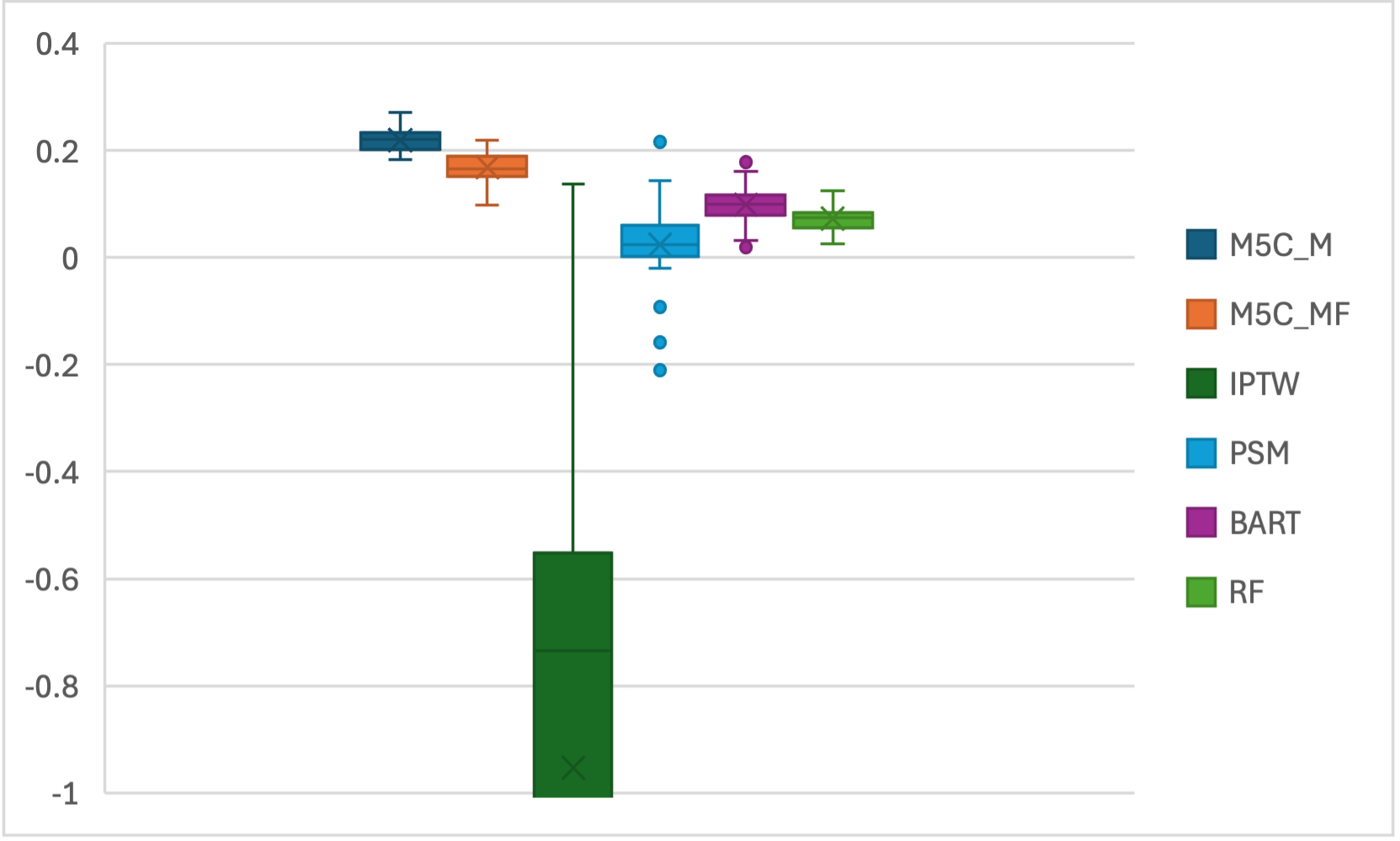}%
        }
    }%
    \hfill
    \subfloat[95\% confidence interval of the ATT estimation]{%
        \resizebox{0.49\textwidth}{!}{
        \includegraphics[width=\textwidth]{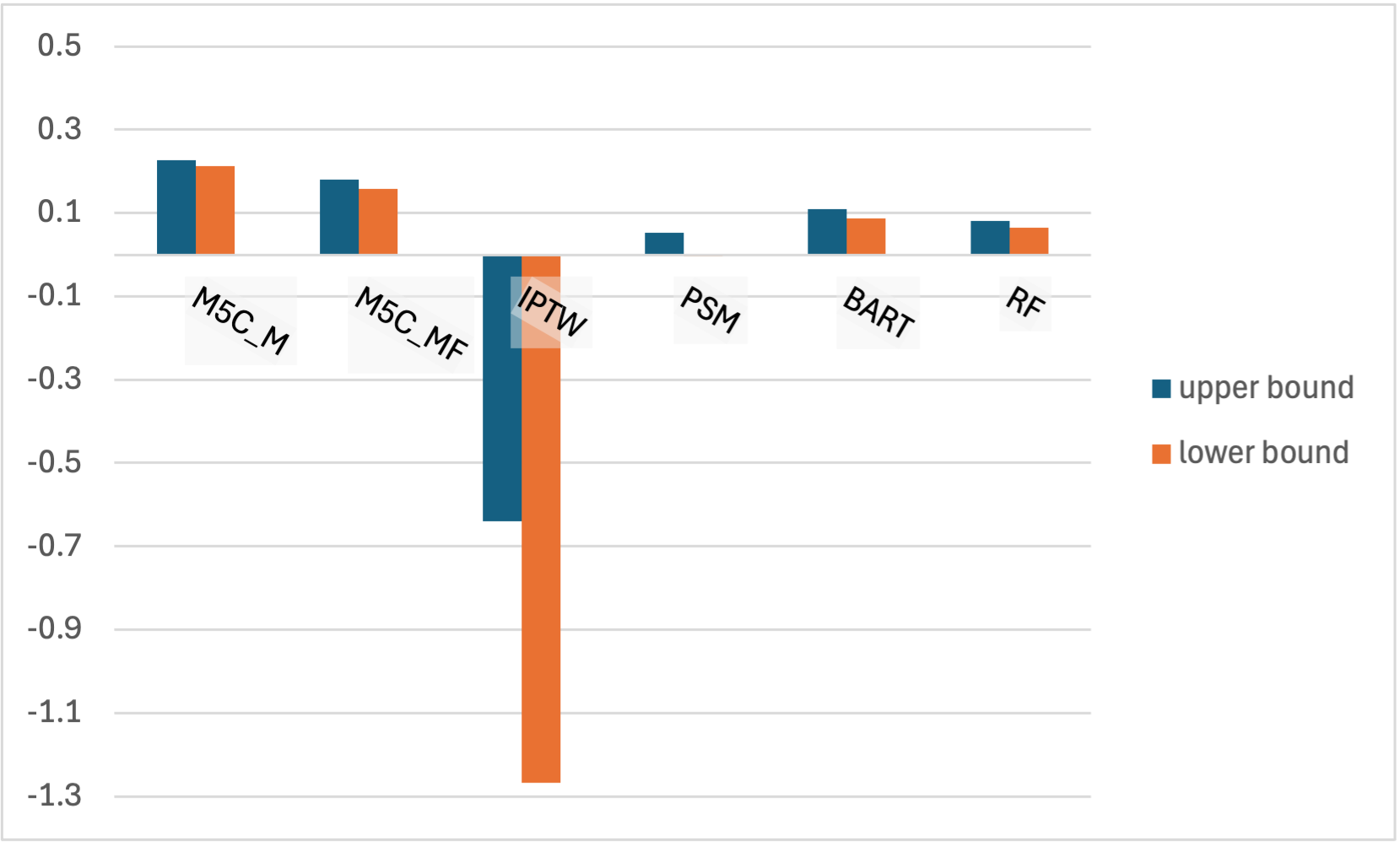}%
        }
    }%
    \hfill
    \caption{ATT estimation for CDC diabetes dataset for each algorithm}
    \label{fig:cdcatt}
\end{figure}

When analyzing this dataset, we employed a bootstrap strategy that randomly selects 500 participants with a treatment value of 1, while keeps the full control dataset. We repeated this process 30 times across the dataset, applying both our proposed method and several benchmark algorithms. For each iteration, we computed the ATT estimation for each method, along with the corresponding confidence interval. The results are presented in Figure \ref{fig:cdcatt}.

According to Figure \ref{fig:cdcatt}, our proposed algorithm consistently estimates a positive and statistically significant ATT, which aligns with the intuition that individuals who perceive themselves to be in poor health (\texttt{GenHlth} = 5) are more likely to have diabetes than other groups. In contrast, in the benchmark algorithms,  propensity score based algorithm frequently produce counterintuitive results. While methods such as BART and Random Forest also yield positive ATT estimates, the magnitudes are relatively small and some values are not significant, potentially making it more difficult for decision-makers to draw meaningful conclusions. 

The average time spent for running each algorithm is displayed in Table \ref{tbl:cdctime}.

\begin{table}[htbp]
\centering
\caption{CPU time for each method to process CDC dataset (in seconds)}
\resizebox{0.8\textwidth}{!}{
\begin{tabular}{||c||c|c|c|c|c|c|}
\hline
\textbf{Model}    & M5C\_M     & M5C\_MF    & IPTW   &  1PSNNM & BART & RF \\ \hline
\textbf{CPU Time} & 44.046 & 117.827 & 0.130  & 0.708  & 165.763 & 67.966 \\ \hline
\end{tabular}
}
\label{tbl:cdctime}
\end{table}

\begin{figure}[htbp]
  \centering
  
  \begin{subfigure}{0.5\textwidth}
    \includegraphics[width=0.48\linewidth]{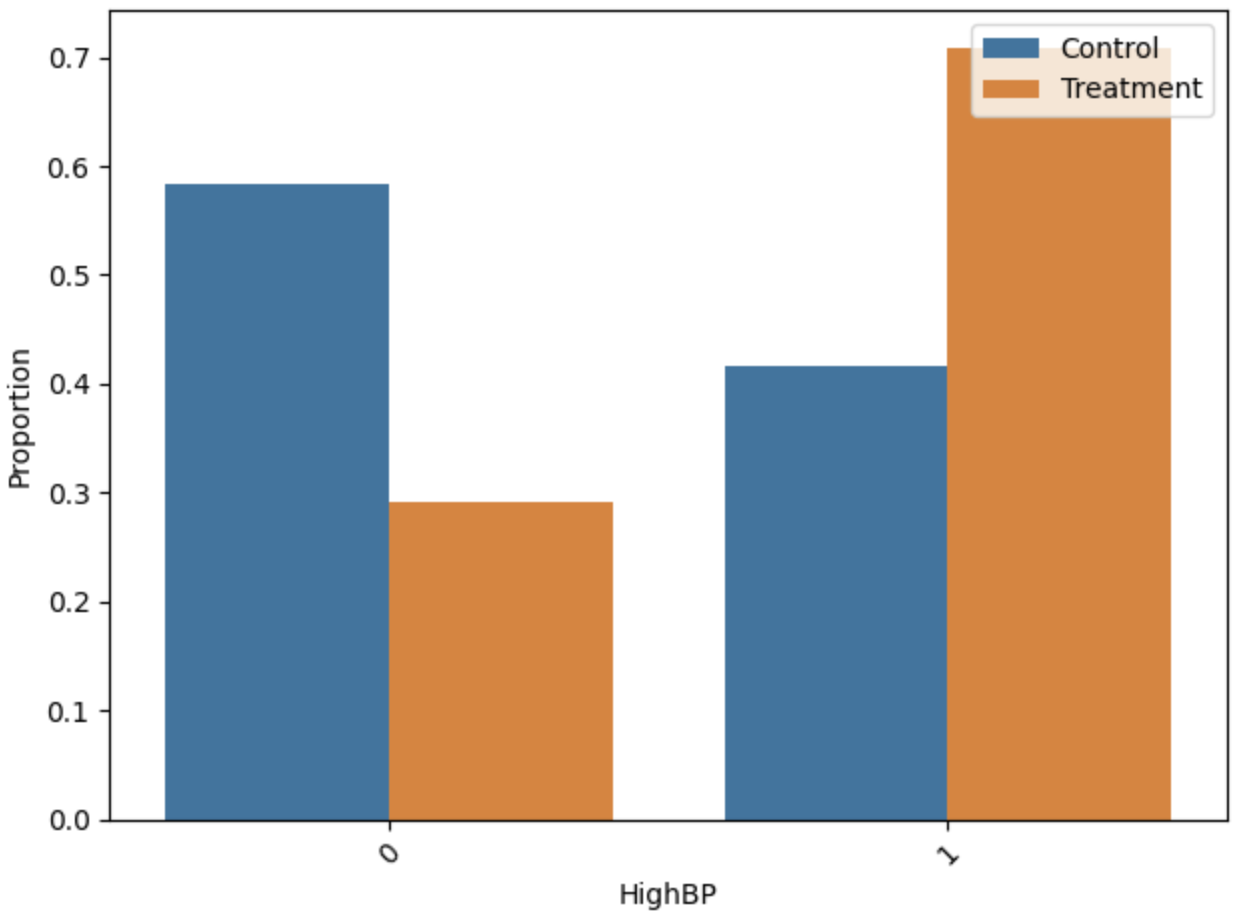}
    \includegraphics[width=0.48\linewidth]{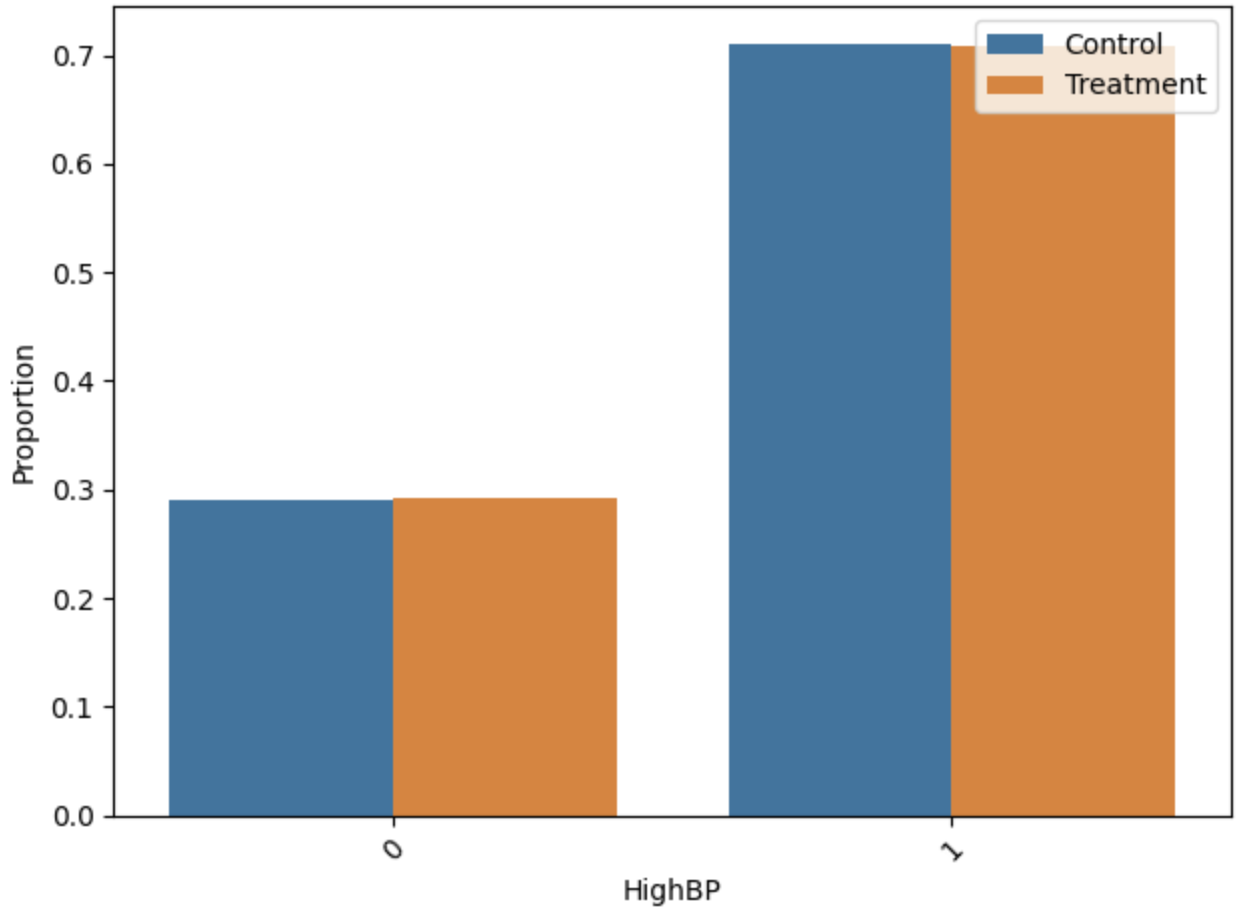}
  \end{subfigure}%
  \hfill
  \begin{subfigure}{0.5\textwidth}
    \includegraphics[width=0.48\linewidth]{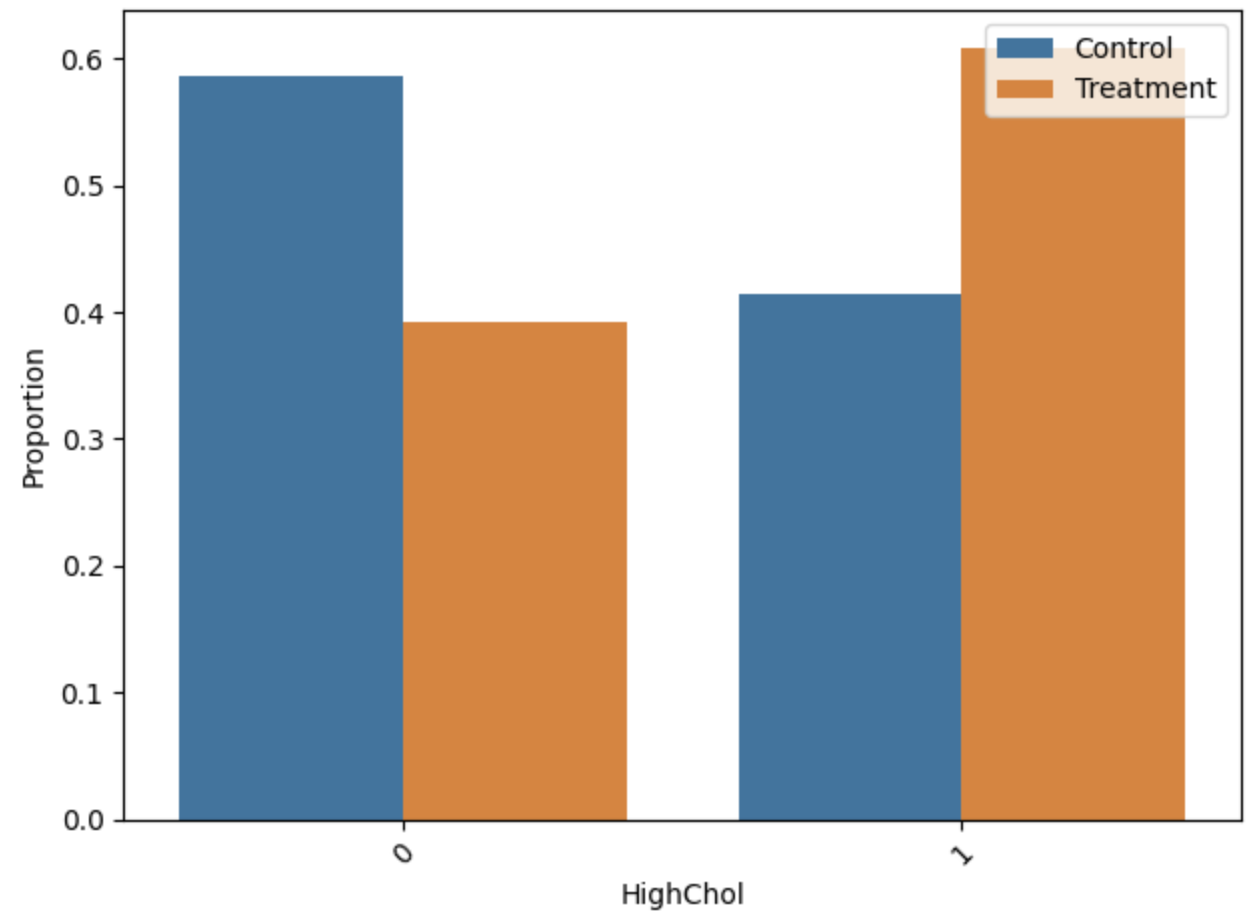}
    \includegraphics[width=0.48\linewidth]{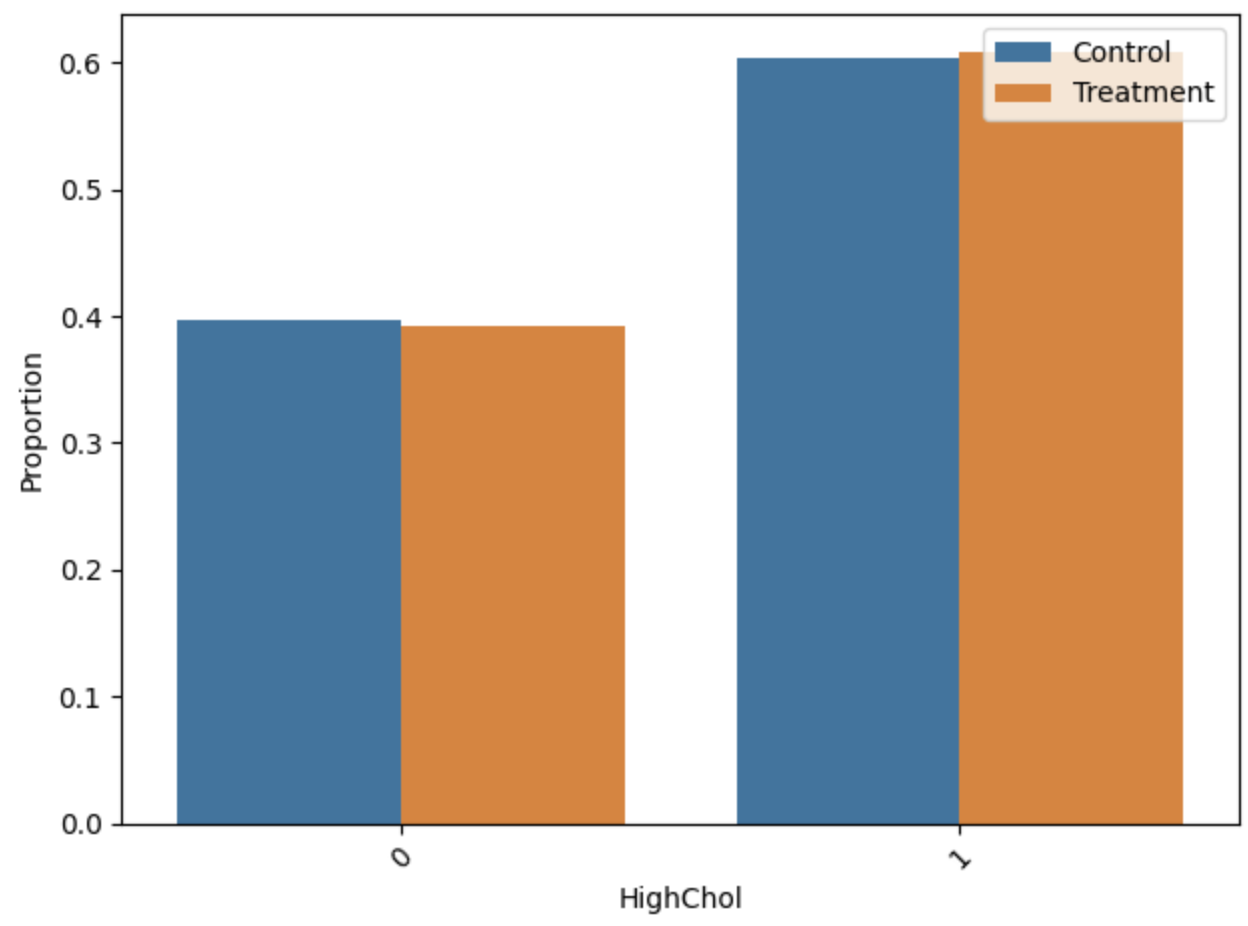}
  \end{subfigure}%
  \hfill
  \begin{subfigure}{0.5\textwidth}
    \includegraphics[width=0.48\linewidth]{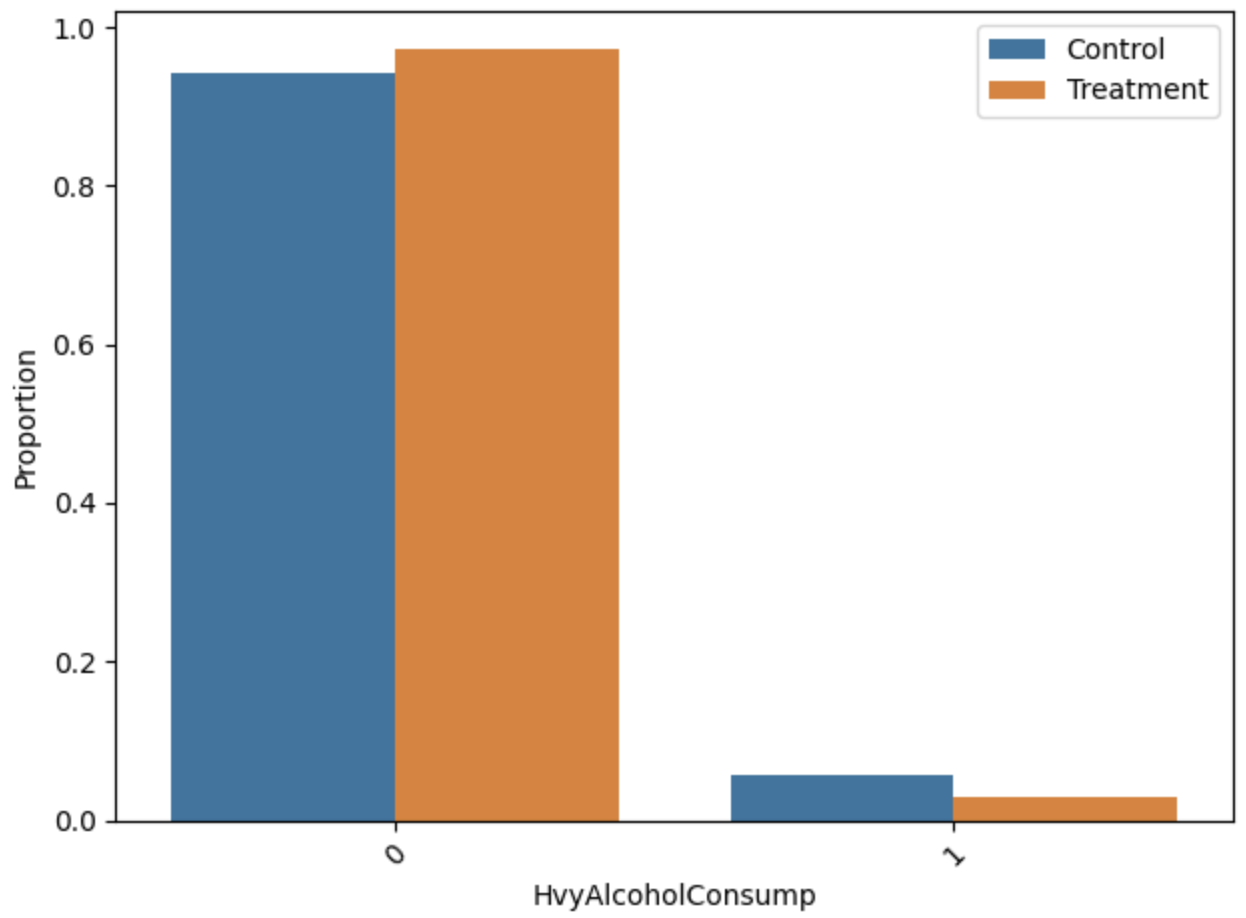}
    \includegraphics[width=0.48\linewidth]{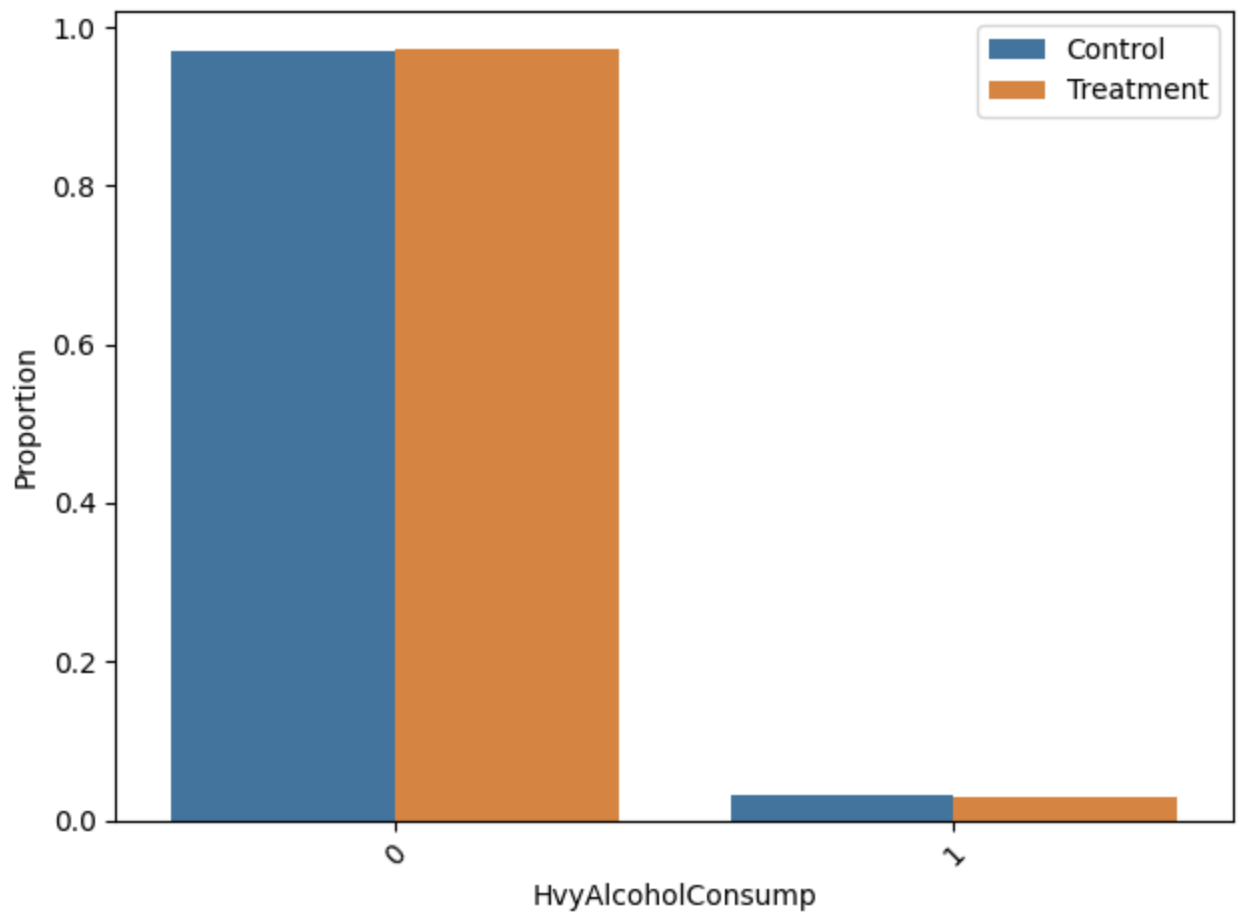}
  \end{subfigure}%
  \caption{Comparison for the distribution of three top important features {\texttt{HighBP} (1 = high, 0 = no high), \texttt{HighChol} (1 = high, 0 = no high), and \texttt{HvyAlcoholConsump} (1 = heavy, 0 = no heavy)} from linear regression before and after matching}
  \label{fig:comp_x}
\end{figure}

We also present the distributions of the top three most important variables (\texttt{HighBP}, \texttt{HighChol}, and \texttt{HvyAlcoholConsump}) before and after matching in Figure \ref{fig:comp_x} (The large graph is presented in \textbf{Online Supplement S1 Appendix F}), based on a coefficients of the logistic regression model fitted to predict the outcome variable \texttt{Diabetes\_binary}. The average standardized mean differences (SMDs) for these three variables after matching are \{0.005420, 0.009771, 0.000208\}, indicating strong balance after matching. The corresponding average variance ratios are \{0.994167, 0.992210, 1.001186\}. These metrics suggest that there is no significant distributional differences regarding the important features between the treatment and control groups. 

\section{Conclusion}

This paper proposes a novel matching algorithm for causal 
inference that combines tree-based discretization with integer linear 
programming. It first utilizes a causal M5 model tree for discretization, with each split improves the adjusted R-square metric for the datasets. The tree is constructed using split rules derived exclusively from the control group, and the technique enables the resulting nodes to approximately satisfy the linearity assumption within the control data. Based on this discretization, the algorithm formulates a new integer programming approach to reduce within-stratum bias, thereby yielding more accurate ATT estimates. Compared to state-of-the-art causal inference methods, the proposed algorithm preserves both interpretability and accuracy. Its strong empirical performance suggests high potential for practical applications, particularly in domains such as healthcare and social media, where robust causal decision-making is essential.

\section*{Acknowledgment}%
Financial support from the National Science Foundation (Award Number: 2047094) is greatly acknowledged.



%
%
%


\bibliographystyle{informs2014} 
\bibliography{bibliography} 


\end{document}